\documentclass[runningheads]{llncs}

\usepackage{eccv}

\usepackage{eccvabbrv}

\usepackage{graphicx}
\usepackage{booktabs}

\usepackage[cjk]{kotex}
\usepackage{pifont}
\usepackage{skak}
\usepackage{textcomp}

\usepackage[table]{xcolor}
\usepackage{graphicx}
\usepackage{wrapfig}
\usepackage{caption}

\usepackage{multirow}
\usepackage{float}
\usepackage{xcolor}
\usepackage{enumitem}

\usepackage[accsupp]{axessibility}  

\usepackage{orcidlink}

\usepackage{hyperref}
\hypersetup{
  colorlinks=true,
  breaklinks=true,
  citecolor=eccvblue,
  urlcolor=eccvblue,
  linkcolor=eccvblue
}

\newcommand\ours{\texttt{REFINED-BIAS}\xspace}
\newcommand\cue{cue-conflict\xspace}
\newcommand\Cue{Cue-conflict\xspace}
\newcommand{\Aref}[1]{Appendix~\ref{#1}}

\begin{document}
\title{On the Reliability of Cue Conflict and Beyond}

\begingroup
\renewcommand\thefootnote{}
\footnotetext{$^{*}$ Equal contributions. $^{\dagger}$ Corresponding authors.}
\endgroup

\author{Pum Jun Kim\inst{1,}$^*$\and
Seung-Ah Lee\inst{1,}$^*$\and 
Seongho Park\inst{2}\and
\\
Dongyoon Han\inst{3,}$^\dagger$\and
Jaejun Yoo\inst{1,}$^\dagger$}

\authorrunning{Kim et al.}

\institute{Ulsan National Institute of Science \& Technology \and
College of Medicine at Hanyang University \and
NAVER AI Lab}

\maketitle

\begin{abstract}
Understanding how neural networks rely on visual cues offers a human-interpretable view of their internal decision processes. The \cue benchmark has been influential in probing shape-texture preference and in motivating the insight that stronger, human-like shape bias is often associated with improved in-domain performance. However, we find that the current stylization-based instantiation can yield unstable and ambiguous bias estimates. Specifically, stylization may not reliably instantiate perceptually valid and separable cues nor control their relative informativeness, ratio-based bias can obscure absolute cue sensitivity, and restricting evaluation to preselected classes can distort model predictions by ignoring the full decision space. Together, these factors can confound preference with cue validity, cue balance, and recognizability artifacts. We introduce \ours, an integrated dataset and evaluation framework for reliable and interpretable shape–texture bias diagnosis. \ours constructs balanced, human- and model- recognizable cue pairs using explicit definitions of shape and texture, and measures cue-specific sensitivity over the full label space via a ranking-based metric, enabling fairer cross-model comparisons. Across diverse training regimes and architectures, \ours enables fairer cross-model comparison, more faithful diagnosis of shape and texture biases, and clearer empirical conclusions, resolving inconsistencies that prior \cue evaluations could not reliably disambiguate. Our code is available at \href{https://pumjunkim.github.io/REFINED-BIAS/}{\ours}.
\end{abstract}

\begin{figure}[t]
\centering
\includegraphics[width=\linewidth]{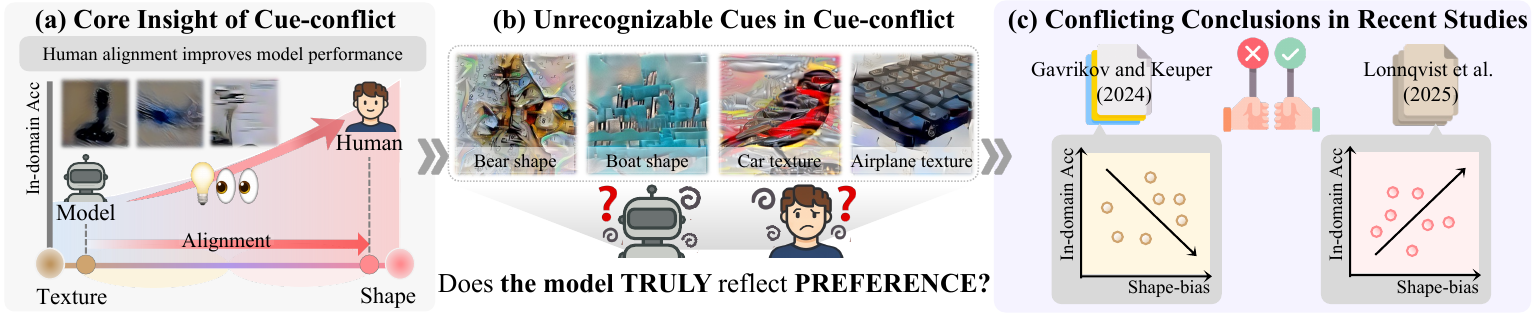}
\vspace{-2em}
\caption{Empirical instability of stylized image-based bias evaluation. (a) illustrates the core insight of \cue that stronger shape bias, similar to humans, improves in-domain performance. (b) shows examples of unrecognizable cues in the \cue dataset. (c) illustrates conflicting findings on this core insight of \cue.}
\label{fig:figure1}
\vspace{-1em}
\end{figure}

\section{Introduction}
\label{sec:intro}
\textit{Do neural networks (NNs) exhibit biases in human-like perception, and can we diagnose them in cognitively meaningful terms?} Addressing these questions is important for bridging machine and human vision, revealing how models internalize visual information and providing a basis for more human-like systems~\cite{baker2018deep, hermann2020origins, geirhos2021partial, ding2022scaling, gavrikov2023extended, finder2024wavelet}. Accordingly, a growing line of work has developed diagnostic benchmarks~\cite{geirhos2018imagenet, hermann2020shapes, mohla2022teaching, wen2023does, doshi2025visual, burgert2025imagenet} that evaluate model biases through a human-perceptual lens. Among them, the \cue benchmark~\cite{geirhos2018imagenet} remains the de facto standard and the most widely adopted framework for bias analysis, consistently used in recent studies~\cite{zhang2019interpreting, hermann2020origins, mummadi2021does, ding2022scaling, oliveira2023connecting, li2023emergence, gavrikov2024can, tu2025toward, zhao2025perceptual, hernandez2025dynamic, jeon2025l1}.

The \cue benchmark probes cue preference using stylized \cue images that combine the shape of one class with the texture of another. It introduced a core insight that has been highly influential: a human-like strong shape bias is associated with higher in-domain performance (\cref{fig:figure1}a). While this idea remains highly valuable, we find that the current stylization-based instantiation yields empirically unstable evaluations. Many \cue images do not clearly convey their shape and texture cue information, making them difficult to recognize for both humans and machines (\cref{fig:figure1}b). Such ambiguity helps explain conflicting conclusions in recent studies on whether shape or texture drives model performance~\cite{oliveira2023connecting, aygun2024enhancing, lonnqvist2025contour, gavrikov2024can} (\cref{fig:figure1}c), and why the benchmark can fail to reflect explicitly shape-inducing training strategies.
\vspace{0.3em}

\noindent These observations raise \textbf{fundamental reliability questions} for stylized \cue evaluation: 
\par\vspace{-0.9em}
\begin{enumerate}[label=Q\arabic*:, leftmargin=2.3em]
\item\textit{Do \cue images instantiate perceptually valid and separable shape and texture cues?}
\item\textit{Does the mixed-cue formulation in \cue ensure controlled and balanced shape and texture cues such that both exert comparable influence?}
\item\textit{Are both shape and texture cues reliably recognizable by both humans and models?}
\end{enumerate}
\vspace{-0.7em}

\begin{figure}[t]
    \centering
    \includegraphics[width=\linewidth]{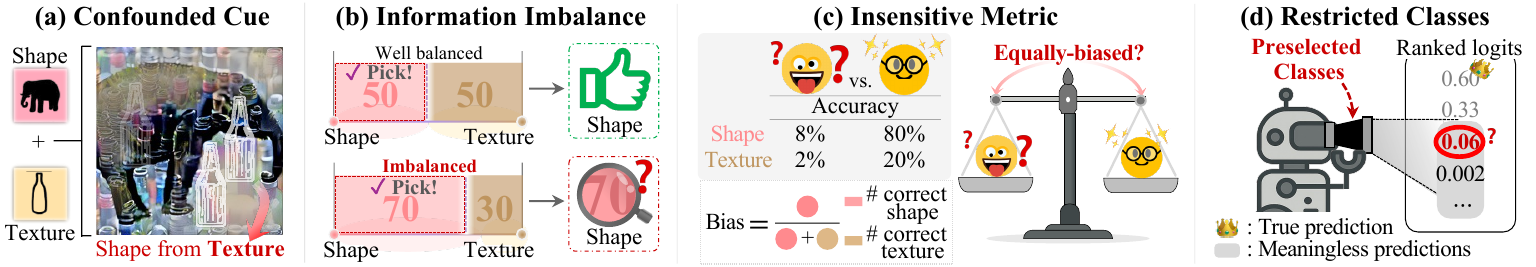}
    \vspace{-2em}
    \caption{Limitations of the \cue benchmark: Although it has offered a valuable and well-designed framework for studying shape and texture biases, we argue that several limitations warrant further attention: 
    (a) cue entanglement caused by stylization, where shape and texture information leak into each other, (b) imbalanced cue information caused by stylization, leading to unfair predictive contributions, (c) ignoring differences in cue sensitivity, which prevents distinguishing models with genuine biases, and (d) evaluation restricted to preselected classes.}
    \label{fig:figure2}
    \vspace{-1em}
\end{figure}

If the answer to any of the above questions is \textit{no}, measured ``preference'' can be confounded by cue construction artifacts rather than reflecting genuine perceptual bias. Here, we identify key limitations of the current \cue instantiation (\cref{fig:figure2}). On the construction side, stylization operationalizes shape and texture through model-dependent features, yielding imperfect cue separation and perceptual impurity (\cref{fig:figure2}a), and provides no control over cue mixing ratios, often producing imbalanced cue informativeness (\cref{fig:figure2}b). On the evaluation side, the ratio-based bias score obscures cue sensitivity, making models with vastly different absolute cue utilization appear similar (\cref{fig:figure2}c), and restricting evaluation to a small subset of labels can distort model predictions, obscuring cue usage in the full decision space (\cref{fig:figure2}d).

To address these issues, we introduce \ours, a reliable framework for integrated evaluation and a disentangled benchmark of interpretable alignment of shape and texture. We define shape as globally and locally coherent geometric structure and texture as scale-consistent repetitive patterns~\cite{hermann2020shapes, ge2022contributions}, and construct balanced, non-overlapping cue pairs from classes that are clearly recognizable to both humans and models. Building on this dataset, we propose a cue-specific sensitivity framework, evaluated on the full label set using a ranking-based metric, enabling preference to be interpreted with absolute sensitivity.

\ours restores diagnostic faithfulness in three ways. (1) It consistently reflects the effects of shape-focused learning strategies on shape-related behavior, whereas \cue exhibits only partial alignment. (2) It separates cue sensitivity from relative preference, allowing fairer cross-model comparison beyond ratio-based bias. (3) It supports more reliable conclusions about how shape and texture sensitivities jointly relate to model performance. As a result, \ours enables clearer empirical insights, including architecture-dependent shape-texture trade-offs, the role of local-to-global mechanisms in improving shape understanding, and bias analysis that extends to both large-scale and quantized vision models, while resolving prior conflicting conclusions that the original \cue benchmark could not reliably disambiguate.

\section{Revisiting Cue Preference Benchmarking}
\label{sec:cue_conflict}
Before introducing our benchmark, we first revisit how cue preference has been measured in prior work and why the widely adopted \cue paradigm requires careful re-examination.
\vspace{-0.5em}

\subsection{The \Cue Paradigm}
The \cue benchmark~\cite{geirhos2018imagenet} measures shape-texture bias using conflicting-cue images that combine the shape of one class with the texture of another through image stylization~\cite{gatys2016image}. Since its introduction, it has become the de facto benchmark for studying cue preference. Importantly, however, this stylization-based design was adopted as a pragmatic solution rather than the only valid way to measure cue preference. In the original study, 
Geirhos \etal~\cite{geirhos2018imagenet} initially considered more purified stimuli, such as sketches, to isolate visual cues more explicitly. Yet these introduced a substantial domain shift for standard CNNs. Stylization was therefore adopted as a practical workaround that preserved more natural image statistics while still inducing cue conflict. 

Under this implementation, shape and texture cues are operationalized via a stylization model: shape is associated with content features, and texture with style statistics such as Gram-matrix matching. Based on 16 ImageNet superclasses~\cite{geirhos2018generalisation}, the benchmark contains 1,280 \cue images constructed from 160 shape-source images and 48 texture-source images. Not all possible combinations of shape and texture are used. Instead, a subset is selected from 7,680 ($160 \times 48$) candidate pairs, and some source images are reused more frequently than others. Shape bias is then computed as $n_s/(n_t+n_s)$, where $n_s$ and $n_t$ denote correctly predicted shape and texture labels, respectively, and similarly for texture. While useful in practice, its current instantiation introduces several limitations that complicate reliable bias interpretation.
\vspace{-0.5em}

\subsection{Limitations of the Current \Cue Instantiation}
\begin{figure}[t]
    \centering
    \includegraphics[width=\linewidth]{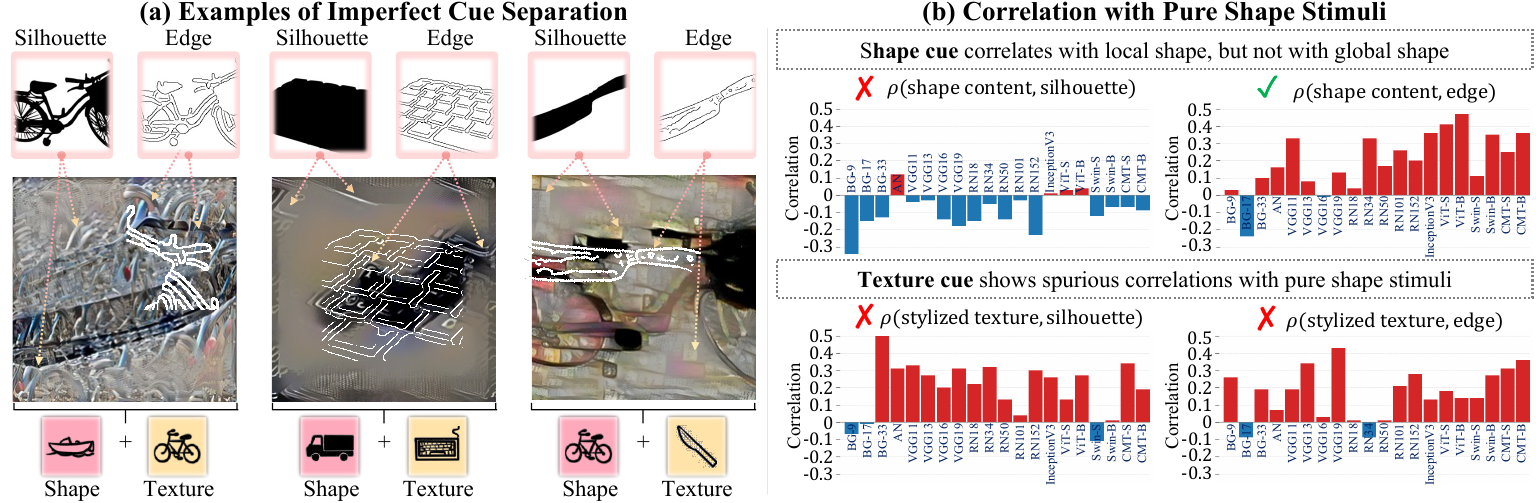}
    \vspace{-2em}
    \caption{Examples of imperfect cue separation in the \cue dataset. (a) Qualitative examples of ambiguously extracted shape and texture cues. (b) Kendall’s rank correlation of class-wise model top-1 accuracy on stylized cues and pure shape stimuli. All ImageNet-1k pretrained CNN and ViT models listed in \Aref{app_sec:model_across_structure} are utilized.}
    \label{fig:figure3}
    \vspace{-1em}
\end{figure}

\subsubsection{Problem 1. Stylization undermines cue reliability.}
\label{sec:problem1}
The \cue presumes that stylization yields two perceptually valid and separable signals: shape (content) and texture (style). However, this operationalization is defined by the stylization model's internal features rather than by human perceptual criteria, and the resulting cues need not align with the intended ``pure'' shape and texture.

We test this assumption by comparing predictions on \cue images against predictions on purified shape stimuli (silhouettes for global shape, edges for local shape). If content is a faithful shape proxy, it should correlate with both silhouette and edge performance, while style should not. Instead, as shown in \cref{fig:figure3}b, content correlates mainly with edge-based signals and fails to track holistic shape, whereas stylized texture exhibits non-trivial correlations with shape proxies, indicating leakage and imperfect disentanglement. Qualitative examples further suggest that shape-like structure can remain visible in texture cues (\cref{fig:figure3}a). Together, these results imply that the current stylization pipeline does not consistently instantiate clean cue separation.

\begin{figure}[t]
    \centering
    \includegraphics[width=\linewidth]{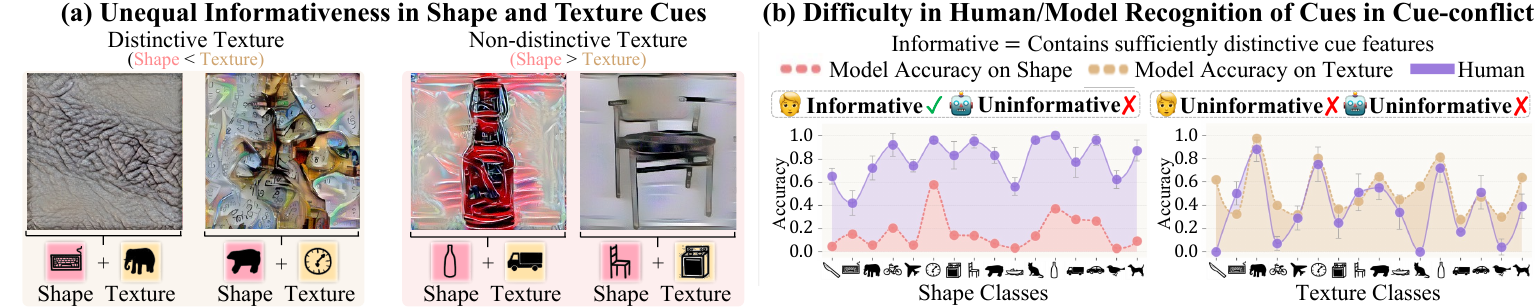}
    \vspace{-2em}
    \caption{Unequal recognizability of cues in the \cue dataset. (a) Qualitative examples of unequal informativeness, and (b) human and model perception trends, on shape and texture cues. Model top-1 accuracies are shown with bars indicating 95\% confidence intervals. All ImageNet-1k pretrained CNN and ViT models listed in \Aref{app_sec:model_across_structure} are utilized. For details on the human study, see \cref{sec:shape_texture_stimuli}.}
    \label{fig:figure4}
    \vspace{-1em}
\end{figure}

Even if such disentanglement were approximately valid, preference measurement requires shape and texture cues to be mixed in an equal ratio (50:50) to measure preferences fairly~\cite{tartaglini2022developmentally, burgert2025imagenet}. Otherwise, the resulting bias score may reflect cue imbalance rather than genuine preference. Yet, stylization offers no explicit control over the relative contribution of shape and texture, and in practice, one cue often dominates the other. As shown in \cref{fig:figure4}a, a keyboard shape blended with an elephant texture contains more texture than shape, making the shape unrecognizable. Similarly, a bear shape mixed with a clock texture is not visible.

These two issues jointly lead to a third problem: the resulting \cue images are often difficult even for humans to recognize in terms of both source cues. If both cues were represented faithfully, humans and models should be able to identify both the underlying shape and the transferred texture with reasonably high accuracy. Instead, as shown in \cref{fig:figure3}a and \cref{fig:figure4} (see also \Aref{app_sec:confounded_cues} and \ref{app_sec:imbalanced_cue}), both human and model recognition are substantially imbalanced, with some texture classes being consistently difficult to distinguish and only a small subset remaining highly recognizable. 
Collectively, the current construction can confound preference with cue validity, cue balance, and recognizability artifacts.
\par\vspace{-1em}

\subsubsection{Problem 2. Relative bias obscures cue sensitivity.}
\label{sec:problem2}
\Cue reports bias as a relative ratio between correct shape and texture predictions. While such a ratio can indicate directional preference, it does not capture how strongly a model actually uses either cue. For example, a model with 8\% shape accuracy and 2\% texture accuracy yields the same relative ratio as a model with 80\% and 20\%, despite the latter being far more sensitive to both cues. Thus, the relative metric conflates directional preference with absolute cue sensitivity. 

This limitation is particularly problematic when the metric is used to compare models or track development progress. An increase in shape-bias score does not necessarily mean that the model has become more shape-sensitive; it may simply indicate that texture sensitivity has deteriorated more severely. Therefore, relative preference can be informative only when interpreted alongside cue-specific sensitivity, rather than as a standalone measure of cue utilization.
\vspace{-1em}

\subsubsection{Problem 3. Restricted label evaluation distorts model predictions.}
\label{sec:problem3}
A further limitation arises from evaluating predictions in a restricted label space. Reliable analysis should be conducted in the full decision space in which the model was trained and performs inference. However, \cue evaluation considers only a subset of candidate labels, typically the shape and texture classes of the constructed image. This filtering can distort the model's actual prediction.

\begin{wrapfigure}{r}{0.5\textwidth}
    \centering    
    \vspace{-2.3em}
    \includegraphics[width=\linewidth]{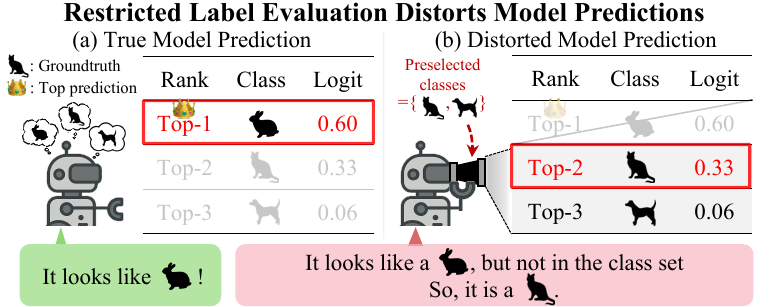}
    \vspace{-2em}
    \caption{
    Illustration of the difference between (a) true model prediction and (b) distorted model prediction. Groundtruth label is ``cat'', true model prediction (\ie, top-1 prediction in full decision space) is ``rabbit''. 
    See \Aref{app:false_positive} for more details.
    }
    \label{fig:figure5}
    \vspace{-2em}
\end{wrapfigure}
    
For example, if a model's top-1, top-2, and top-3 predictions are ``rabbit'', ``cat'', and ``dog'', respectively, the true model prediction is ``rabbit'' (\cref{fig:figure5}a). However, when the model's output space is restricted to a predefined subset of classes (\eg, ``cat'' and ``dog''), the original top-2 prediction (\ie, ``cat'') becomes the top-1 result, thereby distorting the model's true prediction. Such distortion can mislead bias evaluation: when the model's distorted prediction happens to match the ground-truth label (``cat'' in \cref{fig:figure5}b), the model may appear to correctly rely on a given shape or texture cue, even though it did not genuinely recognize it. Consequently, restricted label evaluation can overestimate cue usage and obscure the model's true perceptual behavior.
\vspace{-0.5em}

\subsection{Prior Critiques and the Remaining Gaps}
Our critique is not the first to question aspects of the \cue benchmark. Several recent studies~\cite{tartaglini2022developmentally, burgert2025imagenet, wen2023does, doshi2025visual} have already pointed out important limitations in the current paradigm. Some focused on cue imbalance and attempted to construct more controlled stimuli. For example, Tartaglini \etal~\cite{tartaglini2022developmentally} fill shape silhouettes with textures, but their design overlooks local shape features and still relies on relative bias metrics, limiting fair cross-model comparison. Burgert \etal~\cite{burgert2025imagenet} analyze prediction shifts under cue degradation, but do not fully account for low-level texture cues and remain constrained by ratio-based bias measures. Other studies examined the limitations of relative bias metrics or evaluated cue sensitivity through degraded or texture-suppressed inputs. Wen \etal~\cite{wen2023does} measure sensitivity changes under global shape degradation, whereas Doshi \etal~\cite{doshi2025visual} assess shape sensitivity using texture-suppressed global shape cues.

These studies provide valuable evidence that the current \cue setup should not be treated as an unquestionable standard. At the same time, prior efforts remain largely fragmented, addressing only part of the problem, such as cue balance, relative bias, or shape sensitivity, while leaving other issues unresolved. In particular, no prior work provides a unified re-examination of how cues are constructed, how predictions are evaluated, and how cue-specific sensitivity should be measured in a full-label setting. Our work is motivated by precisely this gap: rather than addressing a single limitation in isolation, we revisit cue preference benchmarking from the ground up and develop an integrated framework that improves cue construction, evaluation, and interpretability together.

Together, the issues discussed in this section suggest that \cue may not fully satisfy the desiderata of a reliable bias benchmark, which leads to unintuitive evaluations in practice (\cref{sec:learning_strategies} and \ref{sec:model_architecture}). Existing alternatives also remain limited in delivering a fully intuitive assessment of model bias (\Aref{app_sec:benchmark_comparison}).

\section{Methodology}
\label{sec:DESTINY}
We propose \ours, a new framework for accurately and reliably measuring and comparing shape and texture biases. First, we define disentangled stimuli following human perceptual standards rather than model-derived heuristics, ensuring that each cue is pure, equally informative, and clearly recognizable (\cref{sec:shape_texture_stimuli}). Second, we introduce a novel bias metric evaluated in the model's true decision space, capturing both preference and sensitivity (\cref{sec:MRR_metric}). By unifying these refined stimuli and evaluation metrics, \ours satisfies the essential desiderata for a reliable bias assessment, enabling a precise and fair comparison of perceptual capabilities across models.
\vspace{-0.5em}

\subsection{Shape and Texture Cue Construction}
\label{sec:shape_texture_stimuli}
\subsubsection{What constitutes shape and texture in our benchmark?}
\begin{figure*}[t]
    \centering
    \includegraphics[width=\linewidth]{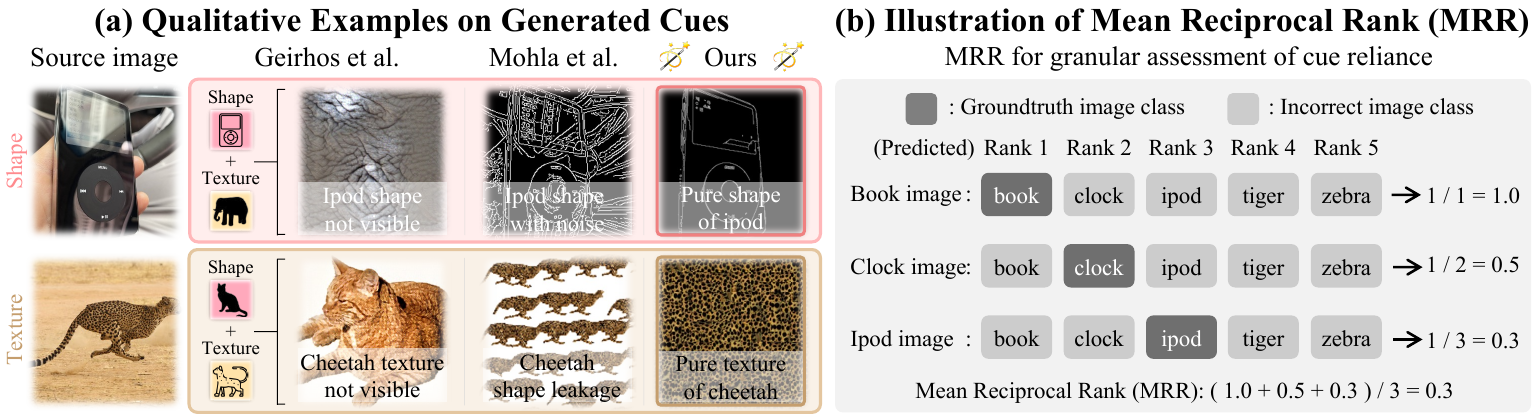}
    \vspace{-2em}
    \caption{
    \ours dataset and metric. (a) Qualitative comparison with existing cue generation methods~\cite{gatys2016image,mohla2022teaching}. (b) Computation of the mean reciprocal rank (MRR) metric. Additional qualitative examples are provided in \Aref{app_sec:category selection}.
    }
    \label{fig:figure6}
    \vspace{-1em}
\end{figure*}

To address \hyperref[sec:problem1]{Problem 1}, we define shape and texture based on human perception rather than model heuristics, and generate cues that faithfully capture these characteristics, as shown in \cref{fig:figure6}a (right).
    
\textit{Texture} is defined as a pattern that consistently repeats within patches of various image sizes. For example, classes like ``honeycomb'' and ``dishrag'' exhibit characteristic textures that remain recognizable even when divided into small patches of different sizes. 
    
\textit{Shape}, on the other hand, is defined as a non-repeating geometric structure, encompassing both global and local features~\cite{hermann2020shapes, ge2022contributions}. Global geometry refers to the overall structure of an object, such as its silhouette, while local geometry includes distinctive substructures not repeated across the object. For instance, although ``ipod'' and ``comic book'' share a similar rectangular global geometry, their local structures are distinct, allowing reliable classification. 

\subsubsection{Data collection.}
\vspace{-1em}
To further mitigate the \hyperref[sec:problem1]{Problem 1}, we construct a curated dataset of 20 ImageNet-derived superclasses (see \Aref{app_sec:category selection} for details), comprising 10 shape-dominant (\eg, clock, hourglass) and 10 texture-dominant (\eg, strawberry, brain coral) categories, selected based on human perceptual judgments. On the other side, the limited source images used in \cue may introduce image-level variations, such as resolution differences that can make patterns appear more coarse or fine. To mitigate this, we collect 300 diverse images per class, 10 from ImageNet and the rest from web sources. This results in a more balanced set of shape and texture cues compared with \cue, which contains only 10 shape and 3 texture sources per class. In total, our dataset contains 6,000 high-quality images, roughly five times larger than \cue.

\subsubsection{Shape and texture cues.}
\vspace{-1.2em}
In designing our cue dataset, we build upon the insights of Mohla \etal~\cite{mohla2022teaching} rather than simply replicating their protocol, as the direct application of existing methods leaves data quality issues unresolved. As shown in \cref{fig:figure6}a (middle), prior methods produce texture cues characterized by local and global shape leakage and reduced cue resolution, while shape cues suffer from noisy background clutter that obscures the object's structural integrity. Furthermore, the stylization method of Geirhos \etal~\cite{geirhos2018imagenet} tends to inherit these inherent data issues found in the \cue benchmark (\cref{fig:figure6}a, left).

To avoid these issues, we design a more precise and carefully controlled cue generation pipeline. As shown in \cref{fig:figure6}a (right), this pipeline ensures clearly recognizable pure shape and texture cues while preserving their full resolution. We further conduct human inspection on each generated cue image to enhance cue quality. The details on the generation pipeline are in \Aref{app_sec:data generation}.
    
\begin{wrapfigure}{r}{0.5\textwidth}
    \centering
    \vspace{-2.2em}
    \includegraphics[width=\linewidth]{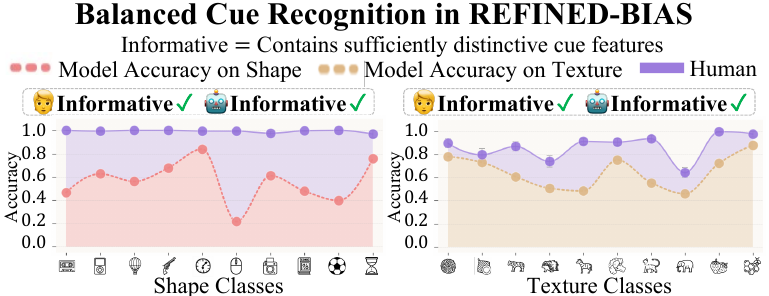}
    \vspace{-2em}
    \caption{Human and model perception trends on shape and texture cues of the \ours dataset. The same experimental setup as in \cref{fig:figure4} is used.}
    \label{fig:figure7}
    \vspace{-2em}
\end{wrapfigure}
    
As shown in \cref{fig:figure7}, our dataset achieves comparable shape and texture accuracies while substantially reducing class imbalance across categories. This reflects the high quality of our stimuli, which also faithfully capture structural 3D information while remaining free from the grid-like artifacts in textures (see \Aref{sec:rb_proxy_corr} for details). Note that the pure stimuli in Geirhos \etal~\cite{geirhos2018imagenet} are small-scale, created for illustration rather than benchmarking, and not fully public. In contrast, our stimuli are systematically constructed and released as a public resource for reliable bias evaluation.

\subsubsection{Mitigating domain shift in CNNs.} 
\vspace{-1em}
To mitigate the domain shift in CNNs that often hinders bias evaluation, we carefully curate a subset of classes where either shape or texture serves as the most discriminative feature for classification. For these categories, we generate clean, artifact-free images to ensure that the intended cues remain clearly recognizable to CNNs (as well as ViTs). Consequently, \ours achieves significantly higher recognition performance across all ImageNet-1k pretrained CNNs (see \Aref{app_sec:model_across_structure}), with average top-1 accuracies of 46\% for shape and 63\% for texture. By comparison, the \cue benchmark achieves 4\% (shape) and 21\% (texture) in full model predictions, and 20\% and 30\% in distorted predictions. This indicates that \ours is significantly less susceptible to the domain shift in CNNs than \cue.
\vspace{-1em}

\begin{wraptable}{r}{0.48\textwidth}
    \vspace{-35pt}
    \centering
    \caption{Inter-rater agreement scores. Fleiss' Kappa ($\kappa$) measures the degree of agreement among multiple raters in their predictions. Higher scores indicate stronger consistency in predictions and reflect how clearly cues are perceived.}
    \label{tbl:kappa}
    \scriptsize
    \centering
    \setlength{\tabcolsep}{6pt}  
    \begin{tabular}{lcc}
    \toprule
    Dataset & Cue Type & Fleiss' Kappa \\ 
    \midrule
    \multirow{2}{*}{\ours}
    & \cellcolor[HTML]{F7CACA} Shape    & 0.9836 \\
    & \cellcolor[HTML]{FFE4B5} Texture  & 0.7973 \\
    \midrule
    \multirow{2}{*}{\Cue}
    & \cellcolor[HTML]{F7CACA} Shape    & 0.7276 \\
    & \cellcolor[HTML]{FFE4B5} Texture  & 0.2937 \\
    \bottomrule
    \end{tabular}
    \vspace{-20pt}
\end{wraptable}

\subsubsection{Our human study details.}
\label{sec:human_study}
To ensure that our shape and texture cues are clearly recognizable to humans, we conducted parallel user studies on \ours and \cue. We recruited 88 participants, each completing 100 classification tasks using randomly sampled images from each benchmark. These tasks were evenly divided into shape and texture sections, where participants identified the target class of a single image for each task. To measure how consistently participants identified these cues, we calculated inter-human agreement using Fleiss' kappa ($\kappa$)~\cite{fleiss1971measuring}. As shown in \cref{tbl:kappa}, while \ours achieved near-perfect agreement for shape ($\kappa=0.98$) and substantial agreement for texture ($\kappa=0.79$), \cue showed significantly lower consistency, particularly for texture ($\kappa=0.29$), indicating inherent ambiguity in its signals. These suggest that \ours provides more recognizable cues to human evaluators.
\vspace{-0.5em}

\subsection{Model Comparisons with Redefined Bias}
\label{sec:MRR_metric}
Addressing the \hyperref[sec:problem2]{Problem 2} and \hyperref[sec:problem3]{Problem 3}, we introduce a new metric that operates on the full logits and enables more fair cross-model comparisons. The core insights behind our metric are as follows: (1) existing bias metrics are often inflated because they rely directly on accuracy for cues, and (2) accuracy appears in both the numerator and denominator, which could distort the ratio. To resolve these issues, we (1) replace accuracy with a ranking-based metric, and (2) apply the metric only in the denominator. While various ranking-based metrics could be used, we employ Mean Reciprocal Rank (MRR)~\cite{voorhees1999trec}, which aligns well with our objectives. While accuracy assigns 0 to both 2nd and 100th place predictions, MRR distinguishes them with values of 1/2 and 1/100, enabling a more precise assessment of how models prioritize different cues.
    
\subsubsection{Redefined Bias.}
\vspace{-1em}
Specifically, our metric computes the reciprocal ranks of the correct shape and texture labels within the model's full prediction ranking (\cref{fig:figure6}b). We refer to these two components as Shape-Sens and Texture-Sens. Unlike conventional MRR, our ranking is computed over the logits:
\vspace{-0.3em} 
\begin{equation}
\text{Shape-Sens}=\frac{1}{N}\sum^N_{i=1}\frac{1}{r_{\text{shape},i}},\quad
\text{Texture-Sens}=\frac{1}{N}\sum^N_{i=1}\frac{1}{r_{\text{texture},i}}.
\end{equation}
\indent Here, $N$ is the total number of samples, $r_{\text{shape},i}$ and $r_{\text{texture},i}$ are the ranks of the correct shape and texture labels for the $i$-th sample in the model's ranked predictions, respectively. The relative bias for shape and texture is defined as:
\par\vspace{-1.2em}
\begin{align}
\text{Shape preference} &= {\text{Shape-Sens}}/({\text{Shape-Sens} + \text{Texture-Sens}}),\\ 
\text{Texture preference} &= {\text{Texture-Sens}}/({\text{Shape-Sens} + \text{Texture-Sens}}).
\end{align}
In the following sections, we demonstrate that our dataset and metric effectively distinguish bias differences and identify models with genuinely strong biases.

\section{Experiments}
Our experimental agenda is guided by two goals. First, we validate the \textit{correctness} of \ours, which can reliably contrast shape-texture bias across a diverse spectrum of training strategies (\cref{sec:learning_strategies}), aligning with our prior understanding and intuition. Building on this foundation, we further investigate how such bias varies across different model architectures (\cref{sec:model_architecture}). We note that the term ``bias'' is used to refer to both preference and sensitivity collectively.
\vspace{-0.5em}

\subsection{Validating \textbf{\ours} Benchmark}
\label{sec:learning_strategies}
The credibility of all subsequent experiments hinges on the correctness of the \ours benchmark; we evaluate whether the outcomes are consistent with our intuition and whether they remain plausible given existing knowledge. To this end, we evaluate the dataset using extensive training strategies for diverse pre-trained models. We then focus more on assessing the correctness of the revised sensitivity metric and examining how the resulting bias measurements correlate with ImageNet top-1 accuracy (\ie, in-domain performance).

\subsubsection{Learning strategies and hypothesis:} 
\vspace{-1em}
We consider 32 ImageNet-1k pretrained models, all based on the same ResNet-50 architecture~\cite{he2016deep}. As a baseline, we use a model trained with random cropping only. Each model applies one additional training strategy on top of this baseline setup. The strategies are as follows (see \Aref{app_sec:model_across_train} and \ref{app_sec:learning_strategies_example} for more details and examples, respectively):
\par\vspace{-0.5em}

\begin{itemize}
\item \textbf{Shape augmentation} explicitly promotes shape-based recognition by exposing models to conflicting cues and enforcing correct shape prediction. We use three models~\cite{geirhos2018imagenet, li2020shape} following this strategy.

\item \textbf{Contrastive learning} implicitly learns cue invariance by aligning representations across texture variations such as blurring, encouraging reliance on stationary shape information. We use three models~\cite{chen2021empirical, chen2020big, caron2021emerging} for this strategy.
        
\item \textbf{Texture distortion} injects noise that disrupts textural information while preserving semantic structure, encouraging reliance on invariant shape features. We use five models~\cite{hendrycks2019augmix, hendrycks2022pixmix, modas2022prime, muller2023classification, hendrycks2021many} following this strategy.

\item \textbf{Mixed augmentation} mixes image pairs or masks regions, 
enabling learning of stable shape and texture cues while reducing reliance on non-stationary cues. We use eight models~\cite{wightman2021resnet}, four with texture distortions~\cite{cubuk2020randaugment}.

\item \textbf{Adversarial training} makes models robust to imperceptible image perturbations~\cite{gavrikov2024can, salman2020adversarially}. It does not directly improve shape or texture perception, so it does not necessarily affect shape or texture bias. We consider 12 models trained with varying levels of noise.
\end{itemize}

\subsubsection{\ours dataset reflects trends clearly.}  
\vspace{-1em}
Based on the hypothesis, we evaluate whether each benchmark dataset faithfully reflects the expected effects of different learning strategies, using the preference metric. As shown in \cref{tbl:table2}, our dataset demonstrates that shape-focused strategies consistently increase shape preference. Notably, even nuanced strategies such as mixed augmentations, which apply mild texture degradation, are accurately reflected by ours as an increased shape preference. While \cue partly captures similar tendencies, many of its results are not statistically significant and show an inconsistent trend across the strategies. 

\begin{wraptable}{r}{0.5\textwidth} 
    \vspace{-33pt}
    \centering
    \caption{$t$-test on the difference between the baseline and training strategies based on the preference. The red, yellow, and gray shading indicate significant 
    {\setlength{\fboxsep}{0pt}\colorbox[HTML]{F7CACA}{shape preference}, \colorbox[HTML]{FFE4B5}{texture preference}, and \colorbox[HTML]{D3D3D3}{no preference}}, respectively ($\alpha$=0.05).}
    \label{tbl:table2}        
    \scriptsize
    \centering
    \setlength{\tabcolsep}{0.3pt}
    \begin{tabular}{lccc}
    \toprule
    Model Family &  Expected & \Cue          &  \ours         \\ 
    \midrule
    {\textcolor[rgb]{1.0000, 0.8431, 0.0000}{\scalebox{1.0}{\ding{108}}}} Mixed Aug   & \cellcolor[HTML]{F7CACA}{shape}  
                & \cellcolor[HTML]{FFE4B5}\textit{p}$=$2.72e-04 
              & \cellcolor[HTML]{F7CACA}\textit{p}$=$0.002    \\
    {\textcolor[rgb]{0.757, 0.490, 0.922}{\scalebox{1.0}{\ding{108}}}} Texture Dist    & \cellcolor[HTML]{F7CACA}{shape}   
                    & \cellcolor[HTML]{F7CACA}\textit{p}$=$0.003
                  & \cellcolor[HTML]{F7CACA}\textit{p}$=$0.010    \\
    {\textcolor[rgb]{0.329, 0.518, 0.922}{\scalebox{1.0}{\ding{108}}}} Shape Aug   & \cellcolor[HTML]{F7CACA}{shape}    
                & \cellcolor[HTML]{D3D3D3}\textit{p}$=$0.181
              & \cellcolor[HTML]{F7CACA}\textit{p}$=$0.018    \\
    {\textcolor[rgb]{0.133, 0.545, 0.133}{\scalebox{1.0}{\ding{108}}}} Contrastive & \cellcolor[HTML]{F7CACA}{shape}   
                    & \cellcolor[HTML]{D3D3D3}\textit{p}$=$0.684
                & \cellcolor[HTML]{F7CACA}\textit{p}$=$0.009    \\
    {\textcolor[rgb]{0.922, 0.537, 0.173}{\scalebox{1.0}{\ding{108}}}} Adversarial  & \cellcolor[HTML]{D3D3D3}{neither}   
                & \cellcolor[HTML]{F7CACA}\textit{p}$=$5.19e-05
              & \cellcolor[HTML]{D3D3D3}\textit{p}$=$0.489 \\ 
    \bottomrule
    \end{tabular}
    \vspace{-20pt}
\end{wraptable}
    
For adversarial training, results on our dataset show that robustness to imperceptible noise does not significantly affect model preference. In contrast, \cue reports a larger increase in shape preference than shape-focused methods, which is counterintuitive since it is primarily aimed at improving adversarial robustness, not shape preference. Furthermore, consistent with recent findings~\cite{burgert2025imagenet} that the ImageNet-1k pretrained ResNet-50 model does not strongly rely on texture cues, our dataset indicates lower reliance on texture (0.49), while \cue reports a higher texture preference (0.77). Overall, these results demonstrate that our dataset provides a more reliable reflection of model behavior.

\subsubsection{Sensitivity metric reveals models that truly utilize cues.} 
\begin{figure*}[t]
    \centering
    \includegraphics[width=\linewidth]{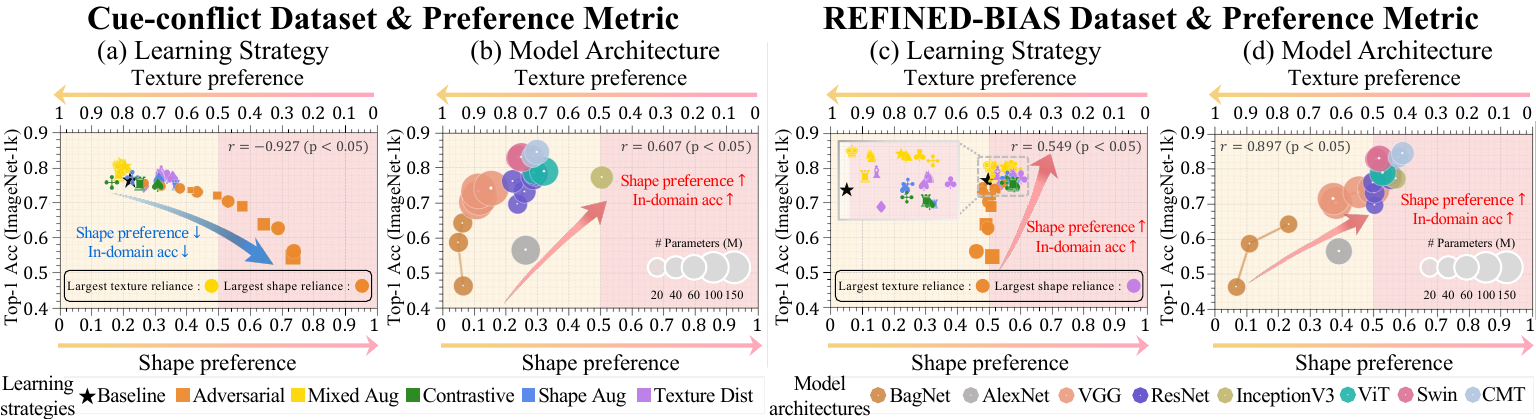}
    \vspace{-2em}
    \caption{
    Comparison of models based on preference metric and ImageNet-1k top-1 accuracy. \textbf{Varying learning strategies with a fixed architecture} are compared on the (a) \cue dataset and (c) our dataset, respectively. \textbf{Varying architectures with fixed learning strategies} are compared on the (b) \cue dataset and (d) our dataset, respectively. Red and yellow backgrounds indicate models with shape and texture preferences, respectively. $r$ is the Pearson correlation coefficient.
    }
    \vspace{-0.5em}
    \label{fig:figure8}
\end{figure*}

\begin{figure*}[t]
    \centering
    \includegraphics[width=\linewidth]{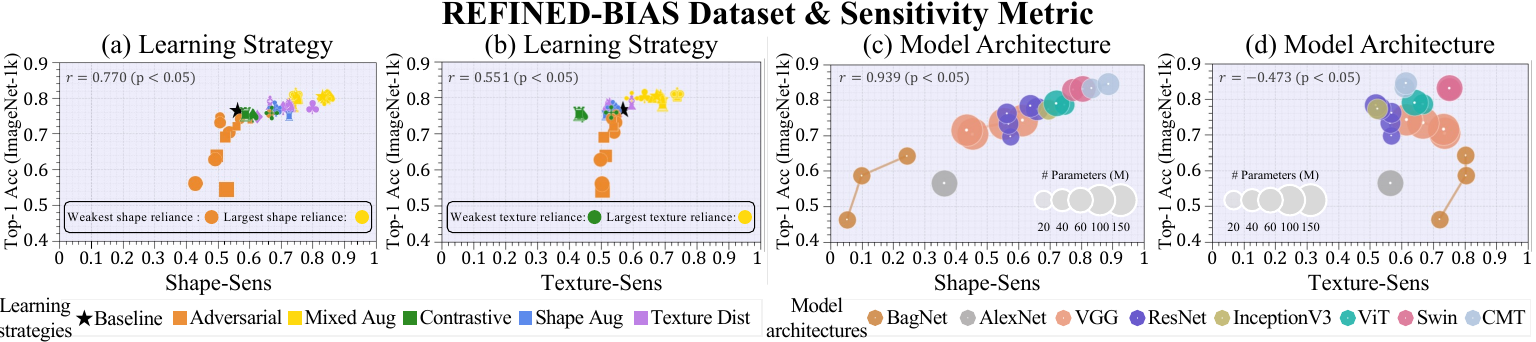}
    \vspace{-2em}
    \caption{
    Comparison of models based on the sensitivity metric and ImageNet-1k top-1 accuracy. (a) shape and (b) texture sensitivity measured across \textbf{different learning strategies with a fixed architecture}. (c) shape and (d) texture sensitivity measured across \textbf{varying architectures with fixed learning strategies}. Correlations are measured via the Pearson correlation coefficient $r$. See \cref{app_sec:trade-off} for additional analysis.
    }
    \vspace{-1em}
    \label{fig:figure9}
\end{figure*}

\vspace{-1em}
The primary goal of our sensitivity metric is to enable fair comparisons across models by reliably distinguishing those that utilize either shape or texture cues. To this end, we evaluate whether it can reveal cross-model differences that the preference metric misses. On our dataset, the preference metric suggests that adversarial learning induces the strongest utilization of texture cues (\cref{fig:figure8}c), while on the \cue dataset, it indicates the strongest utilization of shape cues (\cref{fig:figure8}a). As shown in \cref{fig:figure9}a and \cref{fig:figure9}b, our sensitivity metric demonstrates that adversarial learning does not increase the model's utilization of either cue, whereas mixed augmentations lead models to utilize both shape and texture cues, revealing differences that the preference metric obscures. These results show that our sensitivity metric captures both cross-model differences in cue utilization and independent utilization of shape and texture cues for models that rely on both, unlike the preference metric, which fails to capture either.

\subsubsection{Balanced cue usage positively correlates with performance.} 
\vspace{-1em}
To examine the relationship between model bias and in-domain performance, we employ a dataset with pure, balanced cue information and a sensitivity metric that clearly distinguishes model differences. Following Gavrikov and Keuper~\cite{gavrikov2024can}, we fix the architecture (ResNet-50) to isolate inductive bias as a confounding factor. In \cref{fig:figure9}a and \cref{fig:figure9}b, our benchmark reveals that higher in-domain accuracy positively correlates with the utilization of both shape and texture cues. These results align with prior studies~\cite{li2020shape, mohla2022teaching, tang2026enhancing}, which show that jointly utilizing shape and texture cues improves in-domain performance, further confirming that our benchmark accurately captures their complementary roles.
\vspace{-0.5em}

\subsection{What Becomes Visible Once Bias Is Measured Reliably}
\label{sec:model_architecture}
Based on our dataset and metric, we provide an empirical analysis of how shape and texture biases vary across different model architectures and examine their relation to in-domain performance. See \Aref{app_sec:model_across_structure} for full model details. 

\subsubsection{How ViT design influences cue utilization.}
\vspace{-1em}
The ViT architecture is inherently designed to capture broad, global context through its patch-wise self-attention~\cite{dosovitskiy2020image}, but struggles to effectively encode local features~\cite{yuan2021tokens, peng2021conformer, wu2021cvt}. Subsequent designs like Swin and CMT address this by improving local-to-global feature aggregation: Swin via self-attention in progressively shifted windows, and CMT by combining local convolutional extraction with global self-attention. We investigate how this local-to-global understanding affects model behavior by analyzing shape and texture utilization using our benchmark and sensitivity metric. As shown in \cref{fig:figure9}c, Swin and CMT exhibit higher shape sensitivity than ViT, indicating that improved local feature aggregation enhances shape perception (consistent with Shi~\etal~\cite{shi2023visualization} for Swin). In contrast, \cref{fig:figure8}b shows that the \cue dataset, measured via the preference metric, fails to reveal this advantage: CMT shows little change, and Swin shows reduced shape preference.

\subsubsection{Which drives performance? Shape or texture?}
\vspace{-1em}
Recent studies~\cite{oliveira2023connecting, aygun2024enhancing, gavrikov2024can, lonnqvist2025contour} using \cue benchmark have yielded conflicting conclusions about whether shape preference is more beneficial for in-domain performance. Specifically, as shown in \cref{fig:figure8}a, when the CNN architecture is fixed and only the training strategy varies, higher texture preference is associated with better accuracy, consistent with~\cite{gavrikov2024can}. Conversely, when the training strategy is fixed and the architecture varies, the opposite trend emerges: models with stronger shape preference perform better (\cref{fig:figure8}b), in line with~\cite{oliveira2023connecting, aygun2024enhancing, lonnqvist2025contour}. These contradictory findings within the \cue framework undermine a coherent understanding of preference and call into question whether the benchmark faithfully reflects its intended premise. In contrast, our dataset and preference metric provide consistent results across all setups (\cref{fig:figure8}c and \cref{fig:figure8}d), reliably identifying shape preference as a more important performance contributor. This consistency demonstrates that our benchmark not only accurately captures the intended insight of \cue but also enables a more reliable evaluation of model preference.

\begin{figure}[!t]
\centering
\begin{minipage}[t]{0.49\textwidth}
    \vspace{-11pt}
    \centering
    \captionof{table}{
    $t$-test on the difference between the preference score and the no-preference baseline (\ie, $0.5$) across large-scale vision models. The red and gray shading indicate 
    {\setlength{\fboxsep}{0pt}
    \colorbox[HTML]{F7CACA}{significant} and
    \colorbox[HTML]{D3D3D3}{non-significant}} shape preference, respectively ($\alpha$=0.05).
    }
    \label{tbl:table3}      
    \scriptsize
    \setlength{\tabcolsep}{1.5pt}
    \begin{tabular}{lccc}
    \toprule
    Model & Backbone & \Cue & \ours \\
    \midrule
    \multirow{3}{*}{OpenCLIP}
    & ViT-B/32
    & \cellcolor[HTML]{D3D3D3}{\textit{p}$=$0.276} 
    & \cellcolor[HTML]{F7CACA}{\textit{p}$=$0.002}  \\
    & ViT-B/16
    & \cellcolor[HTML]{F7CACA}{\textit{p}$=$0.037}  
    & \cellcolor[HTML]{F7CACA}{\textit{p}$=$0.002}  \\
    & ViT-L/16
    & \cellcolor[HTML]{D3D3D3}{\textit{p}$=$0.565}    
    & \cellcolor[HTML]{F7CACA}{\textit{p}$=$0.001}  \\
    \midrule
    \multirow{3}{*}{DINOv2}
    & ViT-S/14
    & \cellcolor[HTML]{F7CACA}{\textit{p}$=$0.030}    
    & \cellcolor[HTML]{F7CACA}{\textit{p}$=$3.18e-04}  \\
    & ViT-B/14
    & \cellcolor[HTML]{F7CACA}{\textit{p}$=$0.011}    
    & \cellcolor[HTML]{F7CACA}{\textit{p}$=$0.001}   \\
    & ViT-B/14
    & \cellcolor[HTML]{F7CACA}{\textit{p}$=$5.26e-21}    
    & \cellcolor[HTML]{F7CACA}{\textit{p}$=$3.71e-04}   \\
    \bottomrule
    \end{tabular}
\end{minipage}
\hfill
\begin{minipage}[t]{0.49\textwidth}
    \vspace{0pt}
    \centering
    \includegraphics[width=\linewidth]{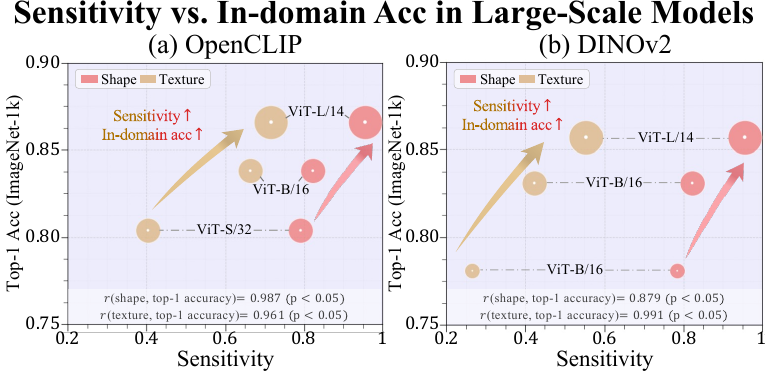}
    \vspace{-2em}
    \captionof{figure}{
    Relationship between cue sensitivity and ImageNet-1k top-1 linear-probe accuracy across large-scale vision models: (a) OpenCLIP, (b) DINOv2. $r$ denotes the Pearson correlation coefficient.
    }
    \label{fig:figure10}
    \end{minipage}
\vspace{-1em}
\end{figure}

\begin{figure*}[t]
    \centering
    \includegraphics[width=\linewidth]{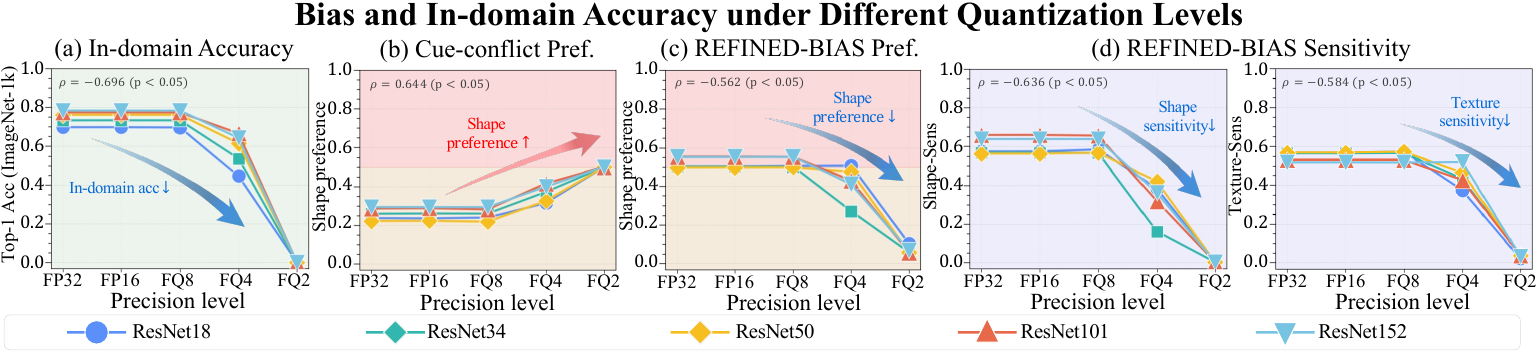}
    \vspace{-2em}
    \caption{
    Bias evaluation on quantized ImageNet-1k pretrained ResNet architectures. Relationship between quantization level and (a) ImageNet-1k top-1 accuracy, (b) preference measured on \cue dataset, (c) preference, and (d) sensitivity measured on our dataset. Kendall's rank correlation $\rho$ is computed and averaged across models.
    }
    \vspace{-1em}
    \label{fig:figure11}
\end{figure*}

\subsubsection{What do biases reveal in large-scale models?}
\vspace{-1em}
Large-scale pretrained models such as CLIP that leverage image-text alignment are known to favor coarse semantics over pixel-level texture, and thereby encourage stronger shape reliance 
~\cite{burgert2025imagenet, hernandez2025dynamic}. This provides a natural testbed for examining two questions: 
Can a benchmark faithfully capture cue preferences for large models induced by different large-scale pretraining strategies, and do these preferences explain downstream recognition performance? To confirm the questions, we evaluate ViT variants from OpenCLIP~\cite{radford2021learning} (pretrained on LAION-2B~\cite{schuhmann2022laion}) and DINOv2~\cite{oquab2023dinov2} (pretrained on LVD-142M), using linear probing on ImageNet-1k with a frozen backbone and a linear classifier. In this setting, \cue fails to reveal a clear shape preference across OpenCLIP models, exposing an evaluation failure mode (\cref{tbl:table3}). In contrast, our benchmark consistently reveals a shape preference, providing a reliable measure of cue preference. Beyond preference, our benchmark further shows that both shape and texture sensitivities jointly drive in-domain accuracy on OpenCLIP (\cref{fig:figure10}a) and DINOv2 (\cref{fig:figure10}b), consistent with our findings in \cref{fig:figure9}. Together, these results demonstrate that our benchmark reliably captures cue bias across diverse large-scale vision models.

\subsubsection{What changes in bias under model quantization?}
\vspace{-1em}
Another practical case that provides a useful testbed is model quantization. Quantization progressively degrades intermediate feature representations, potentially compromising a model's visual cue perception even when semantic predictions remain relatively robust~\cite{nagel2020up}. This raises the question of whether existing cue-bias benchmarks faithfully capture how visual cue processing changes under quantization. To investigate this, we analyze quantized ResNets under precisions from 32-bit to 2-bit. Under stronger quantization, \cue reports an increasing shape preference (\cref{fig:figure11}b), suggesting a shift toward shape reliance. However, because quantization accumulates errors throughout intermediate representations, this trend is difficult to interpret. Moreover, the reported increase in shape preference is accompanied by a decrease in in-domain accuracy (\cref{fig:figure11}a), contradicting \cue's central observation that stronger shape bias is associated with better recognition performance. In contrast, our benchmark shows that both shape and texture sensitivities consistently decrease as quantization becomes stronger (\cref{fig:figure11}d), indicating a gradual degradation of visual cue perception, while shape preference also decreases (\cref{fig:figure11}c), consistent with the drop in in-domain accuracy. These results suggest that our benchmark faithfully tracks changes in model behavior under quantization, whereas \cue does not.

\section{Discussion}
\subsubsection{Does \ours still evaluate cue preference?} 
Yes, but with greater precision. By identifying which information a model can actually leverage when provided in its pure form, we derive a more grounded preference that is not confounded by the failure to recognize competing signals. This approach restores evaluative reliability while preserving the core insight of \cue, enabling a more reliable assessment of how training strategies influence preference.

\subsubsection{Could \ours still be affected by domain shift?} 
\vspace{-1em}
A natural concern is that \ours may face the very domain-shift issue encountered in the \cue setting of Geirhos \etal~\cite{geirhos2018imagenet}. Indeed, stylization-based \cue benchmarks emerged in part because earlier attempts could not avoid substantial distribution mismatch when directly isolating cues. Through cleaner stimulus construction and carefully selected distinctive classes, \ours yields substantially higher top-1 accuracies, as reported in \cref{sec:shape_texture_stimuli}. These results suggest that \ours is less susceptible to severe domain shift than prior stylization-based \cue benchmarks. 

\subsubsection{What exactly is improved by the dataset, the metric, and their combination?}
\vspace{-1em}
Our dataset resolves issues of cue purity, recognizability, and balance. It eliminates the inherent ambiguity of stylization and provides a five-times larger, scalable pool of samples that are easier for both humans and models to interpret. Our sensitivity metric enables full-label evaluation to identify genuine cue utilization. By separating ``how much a model knows'' from ``which cue it prefers'', it uniquely reveals models that genuinely utilize both cues. The integrated benefit arises from their combination, providing the empirical lesson that improved performance stems from dual reliance on both cues and that models with local-to-global attention consistently drive stronger shape utilization. Furthermore, our benchmark enables reliable bias capture in both large-scale and quantized vision models, consistently capturing model bias and its relationship to recognition performance, while faithfully reflecting changes in model behavior.

\subsubsection{Potential limitations.} 
\vspace{-1em}
While our shape cues offer diagnostic clarity, they may not fully capture 3D geometry or viewpoint dependencies. Completely isolating texture from residual shape impressions remains an open challenge, and expanding cue classes will further broaden the scope of bias analysis.

\section{Conclusion}
The \cue benchmark has advanced our understanding of how networks use shape and texture, but its uncontrollable stylization, relative bias that obscures cue sensitivity, and strict class restrictions limit reliable bias analysis. To address these issues, we introduce \ours, a comprehensive framework designed for more controlled and precise bias evaluation. Our refined dataset and metric together provide a unified solution that addresses the limitations of \cue and unresolved issues in recent benchmarks. We establish a more principled and dependable framework for evaluating cue-related biases in modern vision models.

\section*{Acknowledgements}
This work was supported by 
the National Research Foundation of Korea (NRF) grant funded by the Korea government (MSIT) (RS-2022-NR071940, RS-2025-02216916), 
the Institute of Information \& communications Technology Planning \& Evaluation (IITP) grant funded by the Korea government (MSIT) (RS-2020-II201336, the Artificial Intelligence Graduate School Program (UNIST), 
RS-2025-25442149), 
the InnoCORE program of the Ministry of Science and ICT (26-InnoCORE-01). 

\bibliographystyle{splncs04}
\bibliography{main}
\newpage

\setcounter{page}{1}
\setcounter{table}{0}
\renewcommand{\thetable}{\Alph{table}}
\setcounter{figure}{0}
\renewcommand{\thefigure}{\Alph{figure}}

\appendix
\begin{flushleft}
{\Large \textbf{Appendix}}
\par\vspace{0.5cm}
\end{flushleft}
\vspace{-2.5em}

\section{Details on \ours}
\subsection{Our Pipeline for Cue Generation}
\label{app_sec:data generation}
To ensure each cue provides a well-recognizable representation for both humans and models, we develop a precise cue generation pipeline that isolates pure cue information from irrelevant features.
\par\vspace{0.3em}
    
\begin{wrapfigure}{r}{0.48\textwidth}
    \vspace{-24pt}
    \centering
    \includegraphics[width=\linewidth]{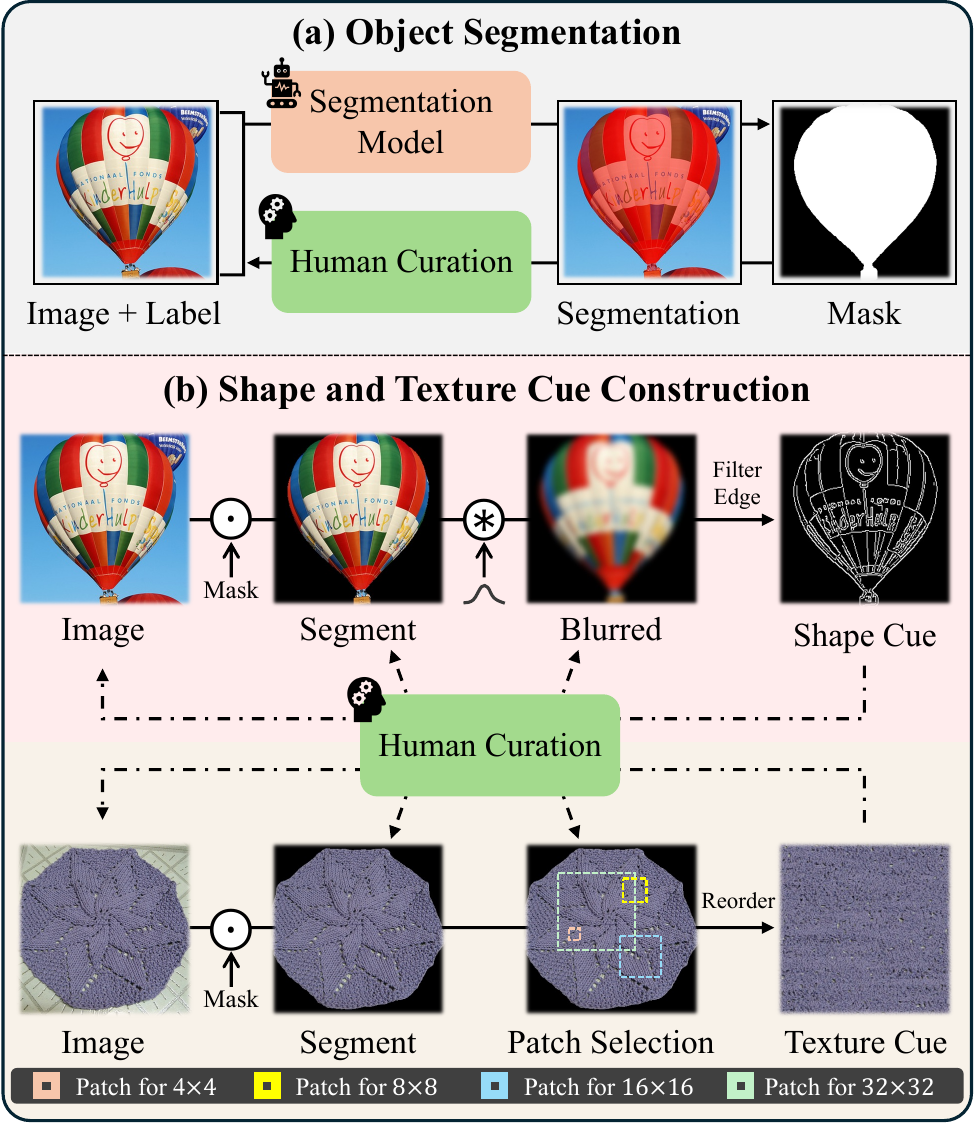}
    \vspace{-2em}
    \caption{Overview of dataset construction for shape and texture cues.}
    \label{app_fig:figure1}
    \vspace{-26pt}
\end{wrapfigure}
    
\noindent\textbf{Shape cue generation.} Our shape cues represent pure structural information by extracting contours exclusively from semantic object regions. We first perform semantic segmentation to isolate the object~\cite{ren2024grounded} and apply a class-adjusted Gaussian blur within the mask to suppress internal texture. From this blurred region, we extract structural contours similar to Mohla \etal~\cite{mohla2022teaching}, rendering them as white edges on a black background. This process preserves both global and local shape features while ensuring the representation remains free from background clutter.
\par\vspace{0.3em}

\noindent\textbf{Texture cue generation.} For our texture cues, the objective is to preserve fine-grained surface patterns while systematically removing any local or global structural information. Building on the semantic segmentation~\cite{ren2024grounded}, we crop patches of four predefined sizes exclusively from the interior of the object. This ensures that the patches do not contain any local contours or boundary information that could inadvertently represent the object's shape. To extract pure texture, these patches are reordered using a strategy similar to that of Mohla \etal~\cite{mohla2022teaching}, eliminating the local structure within the object interior while preventing the formation of grid-like artifacts.
\par\vspace{0.3em}

Finally, we manually curate all generated samples, filtering out instances that fail to represent the intended shape or texture cues, contain noise, or exhibit incorrect semantic masks. 
Note that human curation is not a strict requirement: repeating the validation in \cref{sec:learning_strategies} using our dataset without human curation leaves our conclusions unchanged (\cref{app_fig:rebuttal1}).

\subsection{Category Selection Guided by Visual Cue Distinctiveness}
\label{app_sec:category selection}
Previous benchmarks, such as the 16 superclasses from Geirhos \etal~\cite{geirhos2018imagenet}, were primarily designed to assess model robustness to image degradation. Accordingly, many of the selected classes have large, rigid structures with well-defined silhouettes. These classes are highly recognizable even under strong distortions, making them well-suited for degradation studies. However, they offer little diagnostic value for texture bias analysis, as their surface patterns are often uninformative or homogeneous (\eg, metal surfaces for ``boat'' and ``car''). This leads the texture cues to become less recognizable. To address this, we curated a new set of 20 ImageNet superclasses (10 shape-dominant, 10 texture-dominant) based on human-perceptual criteria:

\begin{figure}[H]
    \vspace{-1.5em}
    \centering
    \includegraphics[width=0.9\linewidth]{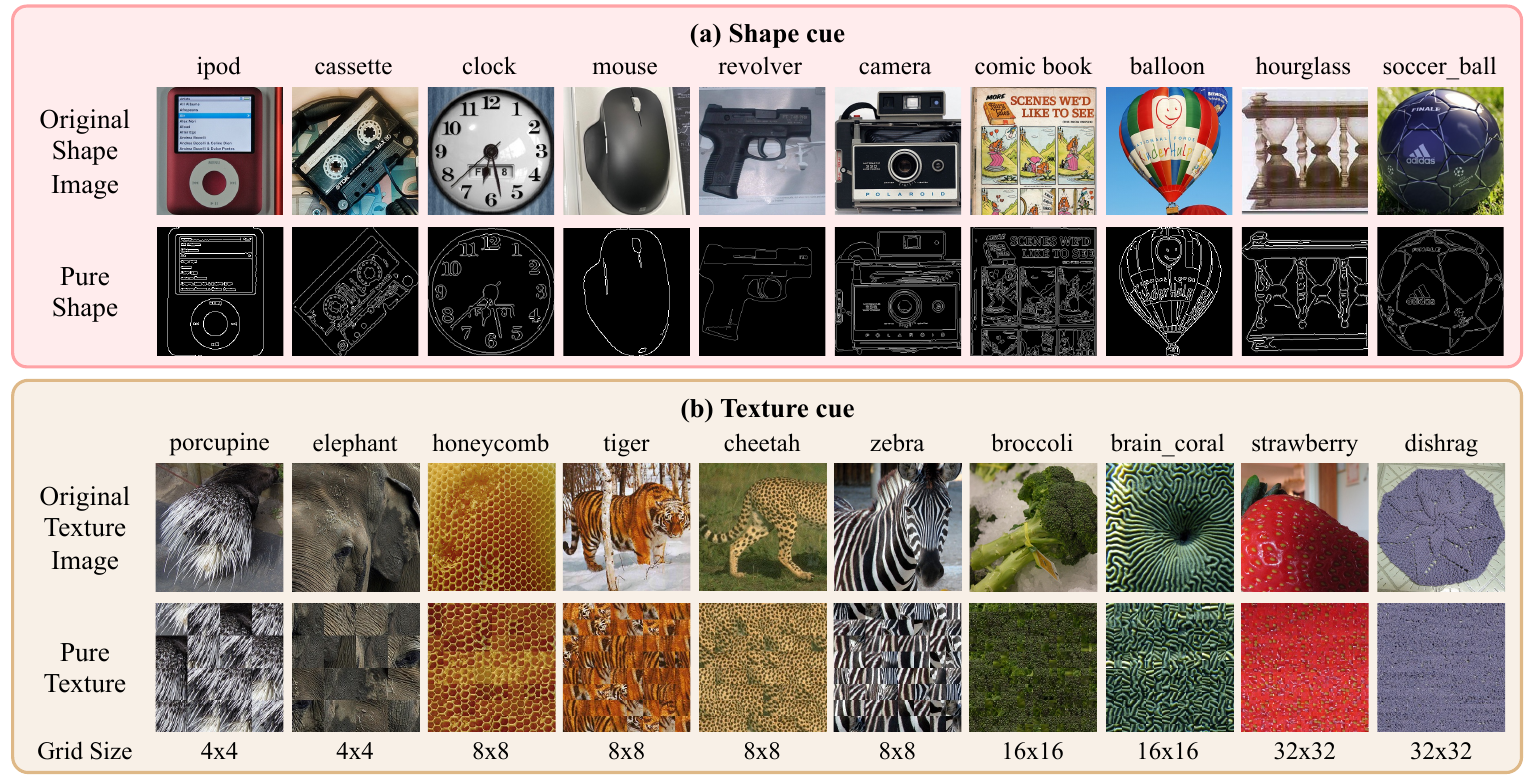}
    \vspace{-0.7em}
    \caption{Examples of shape-texture cues in \ours dataset, along with their corresponding source images for shape and texture.}
    \vspace{-1.5em}
    \label{app_fig:figure2}
\end{figure}

\noindent\textbf{(1) Shape dominant categories.} Shape cue categories were selected for having highly discriminative structural forms with minimal reliance on surface patterns. These include classes such as hourglass and ipod, where both local and global shape are the primary identifying features.
\par\vspace{0.3em}

\noindent\textbf{(2) Texture dominant categories.} Texture cue categories were chosen for featuring rich, class-specific surface textures, with less distinctive structural information. Examples include strawberry, cheetah, and dishrag, where the texture plays a central role in recognition.
\par\vspace{0.3em}

Our current set is intentionally limited to a balanced representation of shape- and texture-dominant categories to ensure both conceptual clarity and practical feasibility. When non-distinctive categories are included, the preference metric yields mixed and statistically insignificant results even under strongly shape-inducing training strategies (as shown in \cref{app_tbl:rebuttal2}), leading to less reliable bias estimates. Distinctiveness is therefore important for reliable bias evaluation.
\newpage

\subsection{Psychophysical User Study}
\label{app_sec:psychophysical_study}
To ensure that \ours provides cues that are clearly interpretable as shape or texture from a human perspective, we conduct a user study. Our web-based survey consists of one hundred questions, evenly divided into two sections that assess human ability to classify images based on shape cues and texture cues. In each question, participants were asked to accurately identify the target class within a given set of classes. For the question set, we randomly sampled an equal number of images from each class for both shape and texture cues. 

Participants first completed the shape-related section, followed by the texture-related section. Before starting each section, participants were shown three randomly selected original images per class to familiarize themselves with the representative shape or texture cues. Following Geirhos \etal~\cite{geirhos2018imagenet}, we inserted pink noise intermittently between questions. Initial participants were randomly recruited from the lab environment, and they were encouraged to share the survey link with people in their personal networks, facilitating broader recruitment through informal social diffusion. The survey was fully anonymous and collected no personally identifiable information. The survey was implemented using Pavlovia, a widely used online platform for psychological experiments. 

For a fair comparison, we conducted a parallel user study on the \cue dataset using the same web-based survey infrastructure. Unlike \ours, where each image is assigned to either the shape or texture section, the \cue study asked participants to label both the shape and texture class for each \cue image. All procedures, such as the use of familiarization examples, pink noise insertion, and anonymous participation, remained identical. This setup ensured a consistent evaluation protocol across datasets, differing only in the nature of the labeling task. A total of 66 raters participated in the \ours study, and 22 raters completed the \cue version.

To compare consistency in participant responses across the two benchmarks, we measured Fleiss' $\kappa$ inter-rater agreement~\cite{fleiss1971measuring}, quantifying how consistently participants labeled each image. On \ours, we observed near-perfect agreement for shape cues ($\kappa=0.98$) and substantial agreement for texture cues ($\kappa=0.79$), suggesting that the cues are reliably recognizable and consistently understood. In contrast, the \cue dataset showed substantial agreement on shape cues ($\kappa=0.72$) but only fair agreement on texture cues ($\kappa=0.29$), indicating that the texture cue in \cue is ambiguous and less interpretable. These findings confirm that \ours provides clearer and more interpretable cues for human evaluators.
\par\vspace{0.5em}
    
\noindent\textbf{Validation of the learning effect.} To validate the presence of any learning effects, we have conducted a reverse-order survey, in which eleven participants completed the texture task first, followed by the shape task. The results remained consistent with those of the original order, with only minor variation (average human shape accuracy: $99\%$ to $96\%$, average texture accuracy: $86\%$ to $91\%$). The slight decrease in shape accuracy, despite potential prior exposure to texture, and the increase in texture accuracy, despite no prior exposure to shape, both counter the presence of learning effects.
\newpage

\begin{figure}[H]
    \centering
    \includegraphics[width=1\linewidth]{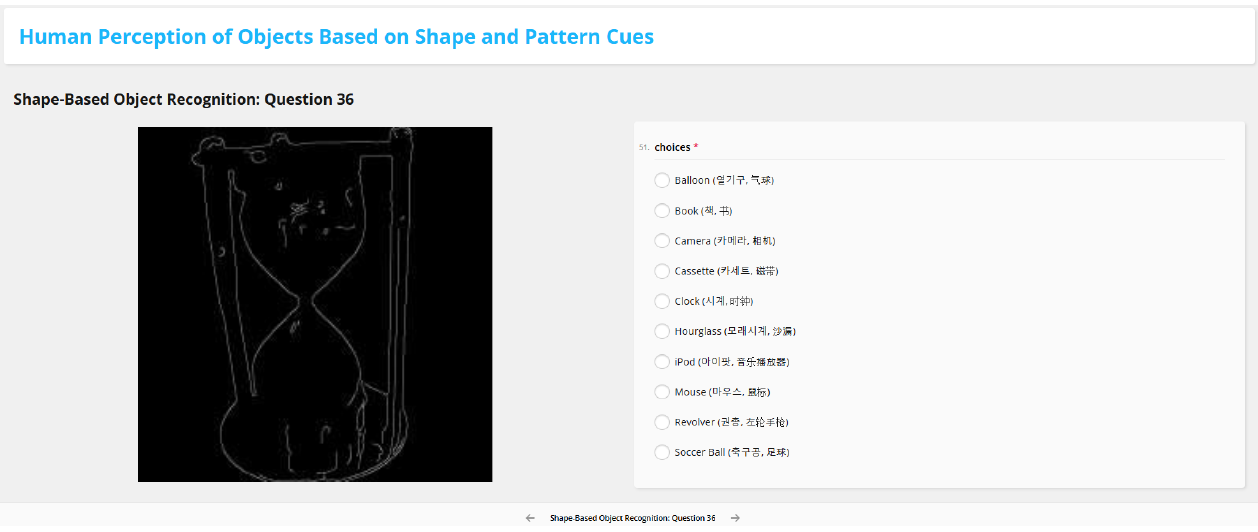}
    \vspace{-1em}
    \caption{Survey example for the shape cue in \ours}
    \label{app_fig:figure3}
    \vspace{1em}
    
    \includegraphics[width=1\linewidth]{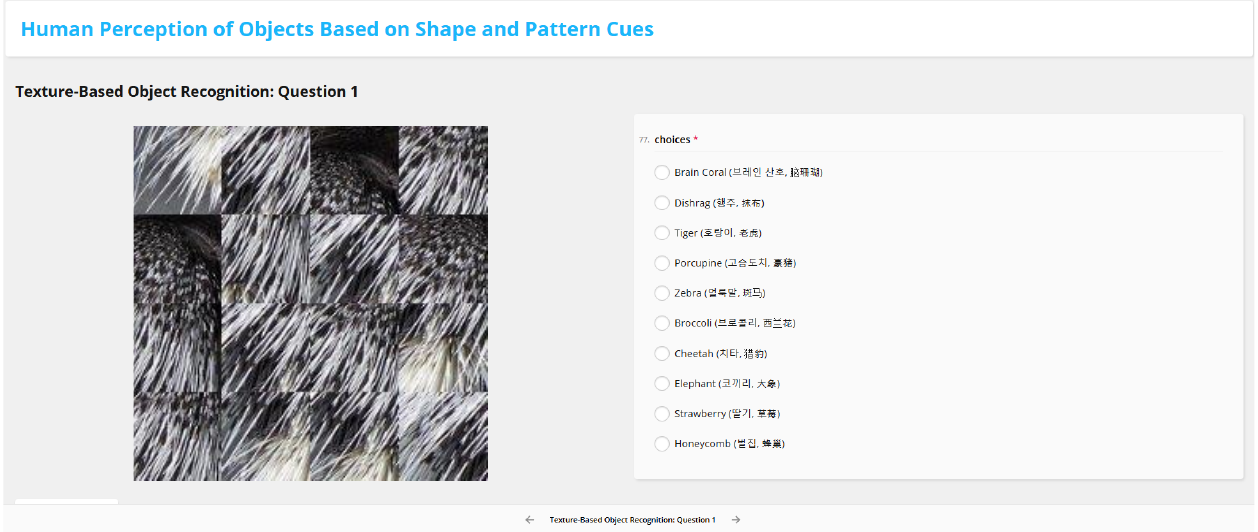}
    \vspace{-1em}
    \caption{Survey example for the texture cue in \ours}
    \label{app_fig:figure4}
    \vspace{1em}
    
    \includegraphics[width=1\linewidth]{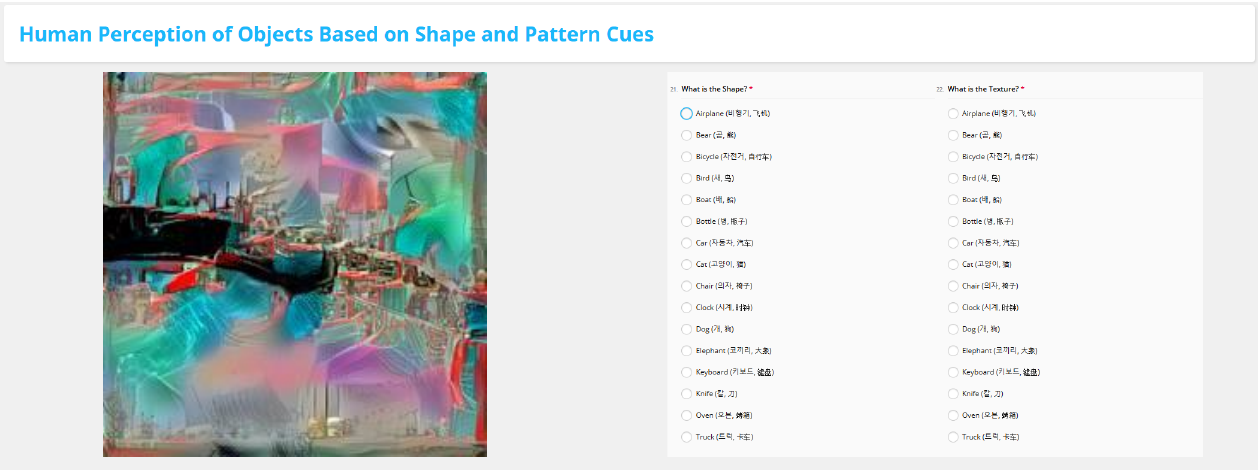}
    \vspace{-1em}
    \caption{Survey example for the shape and texture cue in \cue}
    \label{app_fig:figure5}
\end{figure}
\clearpage

\subsection{Relative Task Difficulty in \ours}
\label{app_sec:metric_compatibility}
When measuring these biases with \ours, the relative task difficulty should also be taken into account. To examine how this is controlled compared to the \cue, we compared the confidence intervals (CIs) between tasks based on human performance shown in \cref{fig:figure3}:
\begin{itemize}
\item \textbf{Avg. accuracy on \ours shape cues:} $0.99$ ($95\%$ CI: $[0.98, 0.99]$)
\item \textbf{Avg. accuracy on \ours texture cues:} $0.87$ ($95\%$ CI: $[0.86, 0.88]$)
\item \textbf{Avg. accuracy on \cue shape cues:} $0.78$ ($95\%$ CI: $[0.76, 0.80]$)
\item \textbf{Avg. accuracy on \cue texture cues:} $0.42$ ($95\%$ CI: $[0.39, 0.46]$)
\end{itemize}

The non-overlapping confidence intervals clearly indicate a difference in difficulty between the two tasks. Importantly, the gap between shape and texture accuracies is smaller in \ours ($0.12$) than in \cue ($0.36$), suggesting that our dataset better controls the relative task difficulty.

\subsection{Bias Computation in \Cue}
\label{app_sec:how_to_compute_bias}
Given stylized images that contain shape and texture labels, the benchmark quantifies the model's bias toward shape or texture based on the proportion of correct predictions aligned with each cue. Specifically, the shape bias is defined as the ratio of correct shape decisions to the total number of correct decisions (number of corrects denoted by $N$):
\[
\text{Shape-bias}=\frac{N_\text{correct-shape}}{N_\text{correct-shape}+N_\text{correct-texture}}.
\]
\indent Similarly, the texture bias is defined analogously as:
\[
\text{Texture-bias}=\frac{N_\text{correct-texture}}{N_\text{correct-shape}+N_\text{correct-texture}}.
\]
\indent A higher shape bias indicates the model relies more on shape than on texture information, whereas a higher texture bias suggests that the model depends more on texture than on shape cues.
\newpage

\section{Qualitative Examples}
\subsection{Examples of Confounded Cues from Stylization}
\label{app_sec:confounded_cues}
\begin{figure}[H]
\centering
\includegraphics[width=1\linewidth]{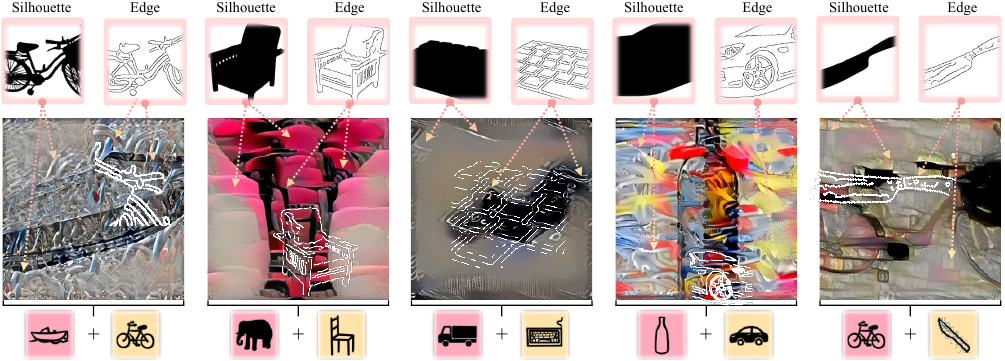}
\caption{Examples of \cue images where \textbf{texture cues are driven by shape information} rather than pure texture. The class markers with red and yellow backgrounds indicate the shape and texture classes, respectively. The white contours are added to highlight the structural information leaked into the texture cue.}
\label{app_fig:figure6}
\end{figure}

\subsection{Examples of Unequally Informative Cues caused by Stylization}
\label{app_sec:imbalanced_cue}
\begin{figure}[H]
\centering
\includegraphics[width=1\linewidth]{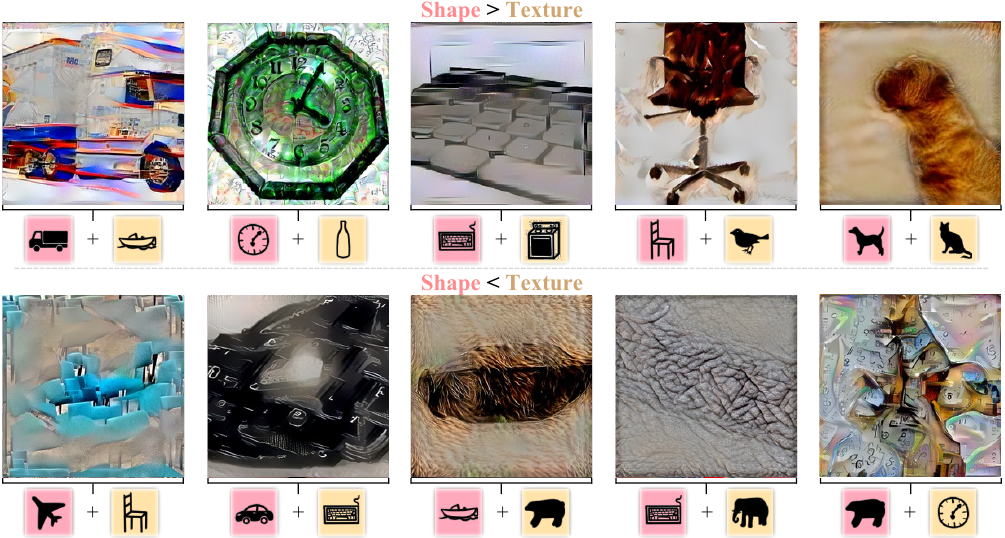}
\caption{Examples of unequally informative cues. The \textbf{top row indicates the shape-dominant} \cue images, while the \textbf{bottom row shows \cue images where texture is dominant}. The class markers with red and yellow backgrounds indicate the shape and texture classes, respectively.}
\label{app_fig:figure7}
\end{figure}
\newpage 

\subsection{Visual Examples for Different Learning Strategies}
\label{app_sec:learning_strategies_example}
\vspace{-1em}
\begin{figure}[H]
    \centering
    \includegraphics[width=0.95\linewidth]{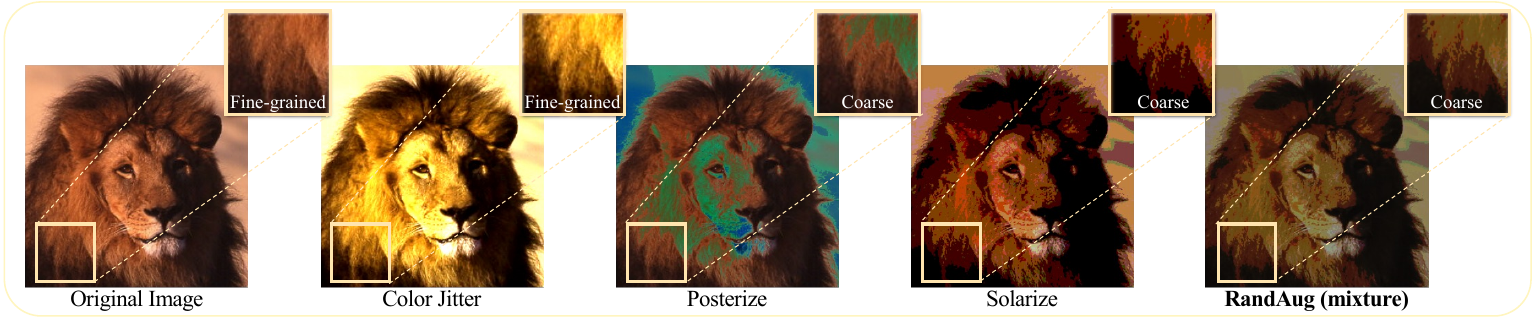}
    \vspace{-1em}
    \caption{Examples of the RandAug method used in mixed-augmentation models. The figure shows the effects of Color Jitter, Posterize, and Solarize augmentations employed by RandAug. The boxed regions in each image highlight the degree of texture change.}
    
    \includegraphics[width=0.95\linewidth]{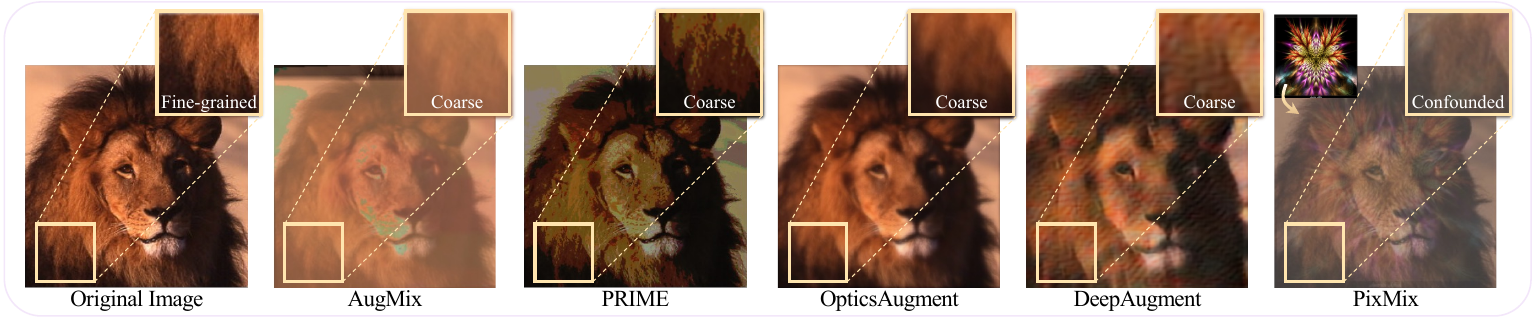}
    \vspace{-1em}
    \caption{Examples of texture degradation methods. The boxed regions in each image highlight the degree of texture change.}
    
    \includegraphics[width=0.95\linewidth]{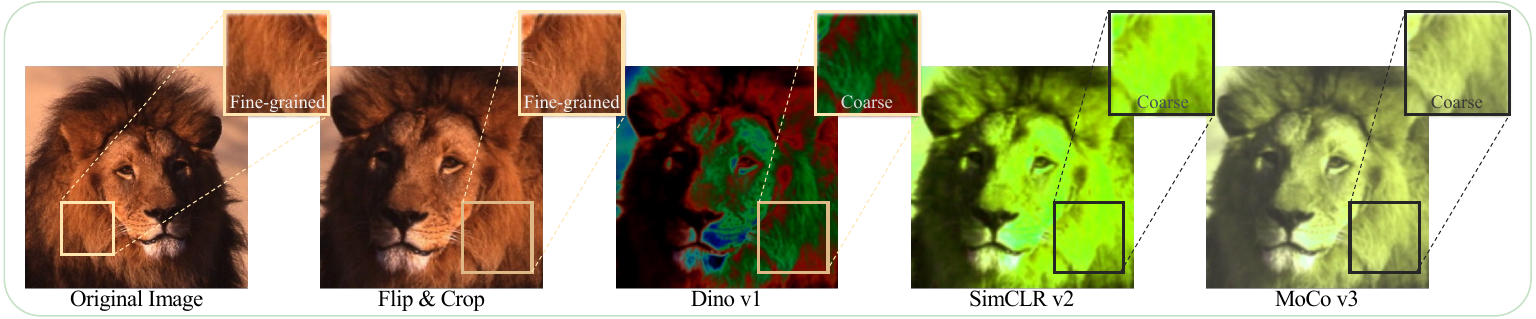}
    \vspace{-1em}
    \caption{Examples of augmentations used for each contrastive learning method. The parameters of all augmentations follow the original model settings. The boxed regions in each image highlight the degree of texture change.}
    
    \includegraphics[width=0.95\linewidth]{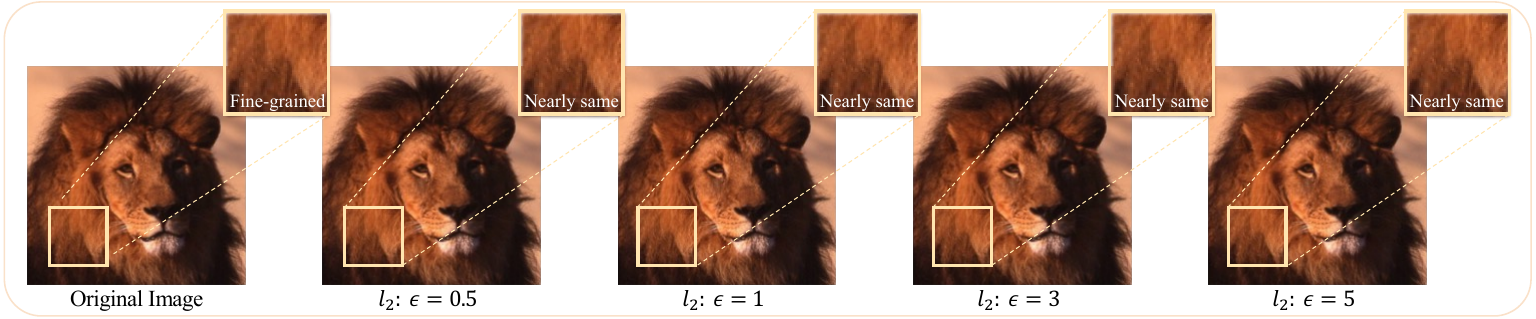}
    \vspace{-1em}
    \caption{Examples of adversarial images generated by adversarial training methods under varying $l_2$ distance. The overall images are perceptually similar in both shape and texture compared to the original images.}
    
    \includegraphics[width=0.95\linewidth]{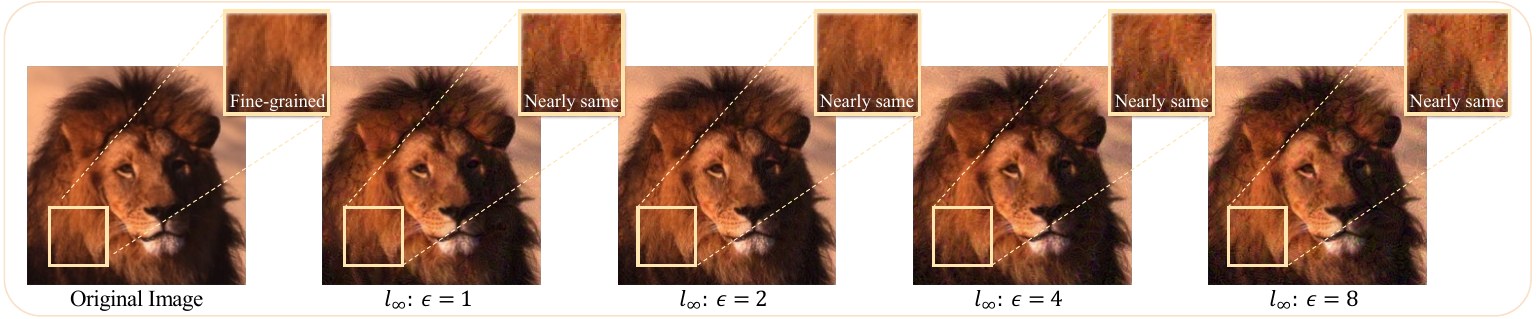}
    \vspace{-1em}
    \caption{Examples of adversarial images generated by adversarial training methods under varying $l_\infty$ distance. The overall images are perceptually similar in both shape and texture compared to the original images.}
\end{figure}
\newpage

\section{Additional Experiments}
\subsection{Distorted Model Prediction in \Cue}
\label{app:false_positive}
\begin{figure}[H]
    \centering
    \vspace{-2em}
    \includegraphics[width=\linewidth]{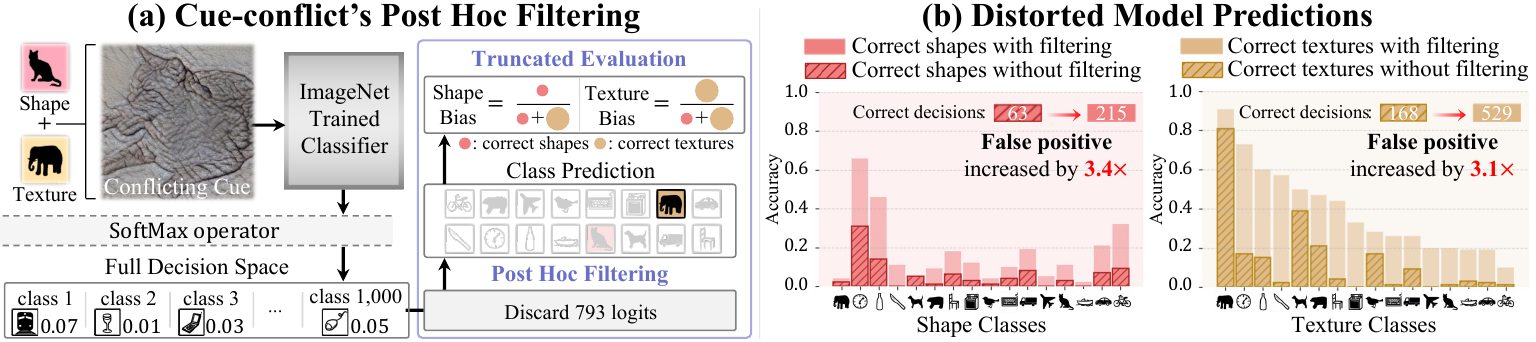}
    \vspace{-0.8em}
    \caption{\textbf{Illustration of post hoc filtering (left)} and its \textbf{induced distortion in model predictions (right)} within the \cue benchmark. Solid and striped bars indicate the average number of correct top-1 predictions across 20 ImageNet-1k pretrained models (all models in \cref{app_tbl:model_arch}) with and without post hoc filtering, respectively.}
\end{figure}

\subsection{Experimental Comparison with Recent Bias Benchmarks}
\label{app_sec:benchmark_comparison}
To experimentally compare \ours with recent benchmarks proposed to address the limitations of \cue, we evaluate whether each benchmark faithfully reflects the expected effects of various learning strategies, following the hypotheses in \cref{sec:learning_strategies}. As shown in \cref{tbl:benchmark_comparison}, \ours consistently reflects an increase in shape preference when shape-focused strategies are applied. In contrast, other benchmarks~\cite{tartaglini2022developmentally, burgert2025imagenet} fail to capture this expected shift. Even under shape augmentation methods, these benchmarks report a texture-preferred result or no preference induction, contradicting the original intent of the training strategies. Note that other recent benchmarks~\cite{wen2023does, doshi2025visual} were excluded from this comparison as their codebases are not publicly available.

\begin{table}[H]
    \vspace{-1.5em}
    \caption{$t$-test on the difference between the baseline and training strategies based on the preference. The red, yellow, and gray shading indicate significant 
    {\setlength{\fboxsep}{0pt}\colorbox[HTML]{F7CACA}{shape preference}, \colorbox[HTML]{FFE4B5}{texture preference}, and \colorbox[HTML]{D3D3D3}{no preference}}, respectively ($\alpha$=0.05).}
    \vspace{-1em}
    \scriptsize
    \centering
    \setlength{\tabcolsep}{3.5pt}  
    \label{tbl:benchmark_comparison}
    \begin{tabular}{lccccc}
    \toprule
    Model Family & Expected & \Cue & Tartaglini \etal~\cite{tartaglini2022developmentally} & Burgert \etal~\cite{burgert2025imagenet} & \ours \\ \midrule
    {\textcolor[rgb]{1, 0.84, 0}{\ding{108}}} Mixed Aug & \cellcolor[HTML]{F7CACA}shape & \cellcolor[HTML]{FFE4B5}\textit{p}$=$2.72e-04 & \cellcolor[HTML]{D3D3D3}\textit{p}$=$0.402 & \cellcolor[HTML]{F7CACA}\textit{p}$=$1.76e-04 & \cellcolor[HTML]{F7CACA}\textit{p}$=$0.002 \\
    {\textcolor[rgb]{0.757, 0.49, 0.922}{\ding{108}}} Texture Dist & \cellcolor[HTML]{F7CACA}shape & \cellcolor[HTML]{F7CACA}\textit{p}$=$0.003 & \cellcolor[HTML]{D3D3D3}\textit{p}$=$0.892 & \cellcolor[HTML]{D3D3D3}\textit{p}$=$0.821 & \cellcolor[HTML]{F7CACA}\textit{p}$=$0.010 \\
    {\textcolor[rgb]{0.329, 0.518, 0.922}{\ding{108}}} Shape Aug & \cellcolor[HTML]{F7CACA}shape & \cellcolor[HTML]{D3D3D3}\textit{p}$=$0.181 & \cellcolor[HTML]{D3D3D3}\textit{p}$=$0.196 & \cellcolor[HTML]{FFE4B5}\textit{p}$=$0.025 & \cellcolor[HTML]{F7CACA}\textit{p}$=$0.018 \\
    {\textcolor[rgb]{0.133, 0.545, 0.133}{\ding{108}}} Contrastive & \cellcolor[HTML]{F7CACA}shape & \cellcolor[HTML]{D3D3D3}p$=$0.684 & \cellcolor[HTML]{D3D3D3}\textit{p}$=$0.146 & \cellcolor[HTML]{D3D3D3}p$=$0.188 & \cellcolor[HTML]{F7CACA}\textit{p}$=$0.009 \\
    {\textcolor[rgb]{0.922, 0.537, 0.173}{\ding{108}}} Adversarial & \cellcolor[HTML]{D3D3D3}neither & \cellcolor[HTML]{F7CACA}\textit{p}$=$5.19e-05 & \cellcolor[HTML]{D3D3D3}\textit{p}$=$0.469 & \cellcolor[HTML]{FFE4B5}\textit{p}$=$2.04e-06 & \cellcolor[HTML]{D3D3D3}\textit{p}$=$0.489 \\ 
    \bottomrule
    \end{tabular}
\end{table}

\subsection{Validity of Shape and Texture Signals in \ours dataset}
\label{sec:rb_proxy_corr}
\begin{figure}[H]
    \centering
    \vspace{-1em}
    \includegraphics[width=\linewidth]{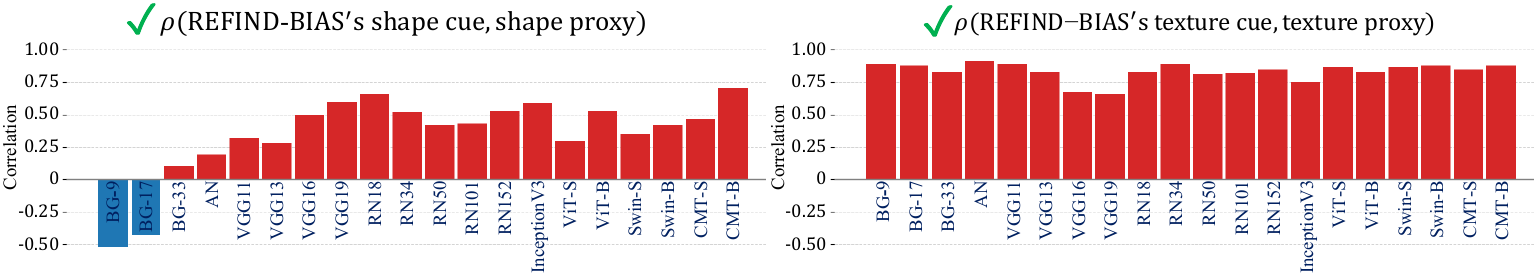}
    \vspace{-2em}
    \caption{Kendall's rank correlation ($\rho$) between class-wise top-1 accuracy on \ours for shape cues and their corresponding shape proxies, and for texture cues and their corresponding texture proxies, respectively.}
    \label{fig:rb_proxy_corr}
\end{figure}

To verify whether \ours's shape and texture cues capture genuine information relevant to shape and texture, we compare them against reliable proxies. Specifically, we extracted depth maps to represent 3D shapes following Yang \etal~\cite{yang2024depth}, and generated textures free from grid artifacts following Efros and Freeman~\cite{efros2023image}. To ensure a comprehensive evaluation, we utilized all 6$k$ source images used in constructing the \ours dataset for this extraction process. These extracted data serve as proxies for valid shape and texture signals, allowing us to evaluate the validity of the information captured by our cues. As in \cref{fig:rb_proxy_corr}, the models' accuracies on these cues positively correlate with their respective proxy accuracies, demonstrating that \ours captures valid shape and texture signals.
    
\subsection{Effect of Human Curation on \ours Dataset}
\begin{figure}[H]
    \vspace{-1.5em}
    \centering
    \includegraphics[width=\linewidth]{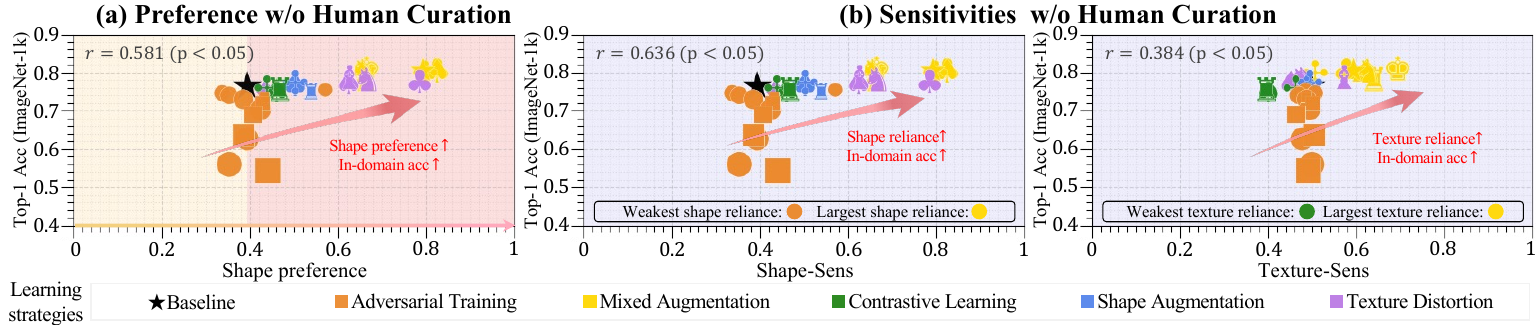}
    \vspace{-2em}
    \caption{
    Effect of human curation on \ours dataset. Comparison of learning strategies under a fixed architecture using (a) the preference metric and our dataset and (b) the sensitivity metric and our dataset, alongside ImageNet-1k top-1 accuracy. (a) reproduces the result in \cref{fig:figure8}c, consistent with the human-curated version; (b) reproduces the results in \cref{fig:figure9}a and \cref{fig:figure9}b, consistent with the human-curated version.
    }
    \label{app_fig:rebuttal1}
\end{figure}

Human curation (\cref{app_fig:figure1}) is introduced to mitigate residual shape/texture leakage and imperfect segmentation, but it is not a strict requirement in the cue generation pipeline. To evaluate whether it can be removed, we regenerate the dataset at the same scale without human curation and repeat the validation experiments in \cref{sec:learning_strategies}. First, we revisit the experiment in \cref{fig:figure8}c, \ie, whether shape preference correlates with in-domain accuracy and whether shape-inducing strategies remain shape-preferring. As shown in \cref{app_fig:rebuttal1}a, the results closely match those of the curated dataset in \cref{fig:figure8}c, while still preserving a significant shape preference ($p<0.05$). Second, \cref{app_fig:rebuttal1}b further shows that the trend where both shape and texture sensitivities contribute to in-domain accuracy remaining consistent with the curated setting (\cref{fig:figure9}a, b). These results suggest that the data generation pipeline can be automated without notable loss in validity, improving scalability and reproducibility while reducing concerns about subjectivity.

\newpage

\subsection{Effect of Ambiguous Categories on \ours Dataset}
\begin{table}[h]
    \vspace{-1.5em}
    \scriptsize
    \centering
    \setlength{\tabcolsep}{3.8pt}
    \renewcommand{\arraystretch}{0.5}
    \caption{
    Effect of ambiguous categories on \ours dataset. $t$-test on the difference between the baseline and training strategies based on the preference. The red, yellow, and gray shading indicate significant {\setlength{\fboxsep}{0pt}\colorbox[HTML]{F7CACA}{shape preference}, \colorbox[HTML]{FFE4B5}{texture preference}, and \colorbox[HTML]{D3D3D3}{no preference}}, respectively ($\alpha$=0.05).
    }
    \vspace{-1em}
    \begin{tabular}{lccc}
    \toprule
    Model Family & Expected & \ours Preference & Preference on Ambiguous Classes\\ \midrule
    {\textcolor[rgb]{0.329, 0.518, 0.922}{\ding{108}}} Shape Augmentation
    & \cellcolor[HTML]{F7CACA}shape
    & \cellcolor[HTML]{F7CACA}\textit{p}$=$0.018 ($<0.05$)
    & \cellcolor[HTML]{D3D3D3}\textit{p}$=$0.919 ($>0.05$)\\
    {\textcolor[rgb]{0.757, 0.49, 0.922}{\ding{108}}} Texture Distortion
    & \cellcolor[HTML]{F7CACA}shape
    & \cellcolor[HTML]{F7CACA}\textit{p}$=$0.010 ($<0.05$)
    & \cellcolor[HTML]{D3D3D3}\textit{p}$=$0.131 ($>0.05$)\\
    {\textcolor[rgb]{0.133, 0.545, 0.133}{\ding{108}}} Contrastive Learning
    & \cellcolor[HTML]{F7CACA}shape
    & \cellcolor[HTML]{F7CACA}\textit{p}$=$0.009 ($<0.05$)
    & \cellcolor[HTML]{D3D3D3}\textit{p}$=$0.707 ($>0.05$)\\
    {\textcolor[rgb]{1, 0.84, 0}{\ding{108}}} Mixed Augmentation
    & \cellcolor[HTML]{F7CACA}shape
    & \cellcolor[HTML]{F7CACA}\textit{p}$=$0.002 ($<0.05$)
    & \cellcolor[HTML]{D3D3D3}\textit{p}$=$0.068 ($>0.05$)\\
    {\textcolor[rgb]{0.922, 0.537, 0.173}{\ding{108}}} Adversarial Training
    & \cellcolor[HTML]{D3D3D3}neither
    & \cellcolor[HTML]{D3D3D3}\textit{p}$=$0.489 ($>0.05$)
    & \cellcolor[HTML]{F7CACA}\textit{p}$=$0.006 ($<0.05$) \\
    \bottomrule
    \end{tabular}
    \label{app_tbl:rebuttal2}
\end{table}
\vspace{-2em}

\subsection{Architecture Dependent Shape and Texture Trade-off}
\label{app_sec:trade-off}
\begin{figure}[H]
    \vspace{-1.5em}
    \centering
    \includegraphics[width=\linewidth]{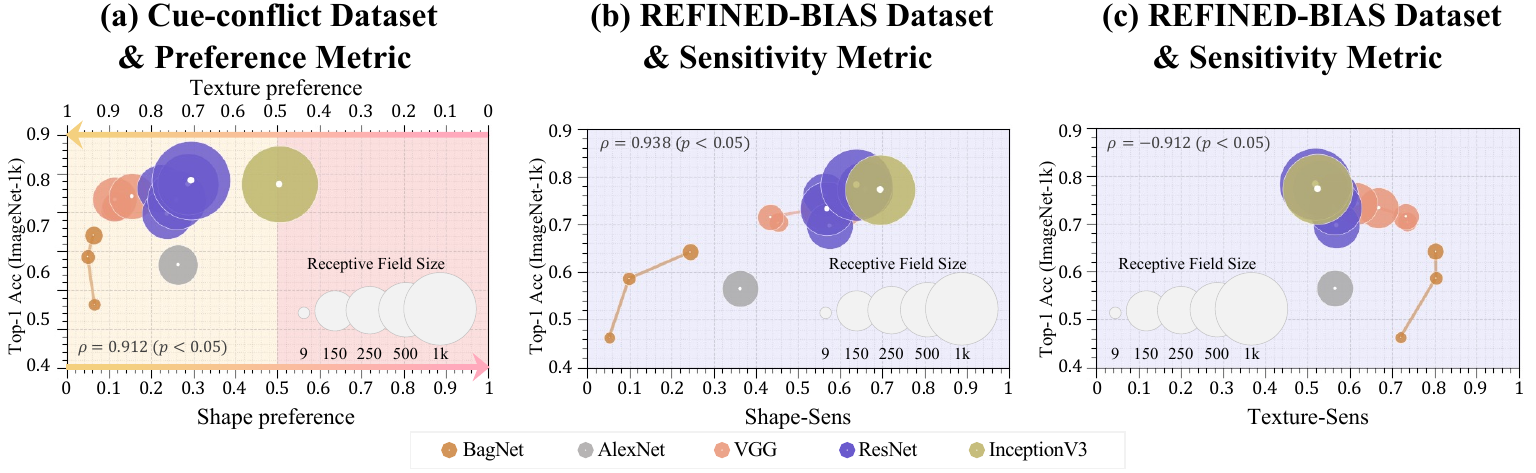}
    \vspace{-2em}
    \caption{
    Comparison of model architectures under a fixed learning strategy and their ImageNet-1k top-1 accuracy. (a) Model preferences are measured on the \cue dataset using the preference metric. (b) Model shape sensitivities and (c) texture sensitivities are measured on the \ours dataset. The point size reflects the theoretical receptive field (RF) size of each model architecture, based on values reported in Araujo \etal~\cite{araujo2019computing}.
    $\rho$ denotes Spearman's correlation coefficient.
    }
    \label{app_fig:trade-off}
    \vspace{-2em}
\end{figure}
We observe an architecture-dependent shape–texture trade-off in the relationship between in-domain accuracy and cue sensitivity (\cref{fig:figure9}c and d). We hypothesize that this behavior can be explained by differences in receptive field (RF) size across CNN architectures. In particular, we observe a positive correlation between RF size and shape sensitivity across CNN architectures (0.938 in \cref{app_fig:trade-off}b). This observation is consistent with prior analyses~\cite{ding2022scaling, liu2022more, zhang2025scaling, cui2024large, miao2025spikingyolox} suggesting that models with larger RF sizes capture broader spatial context, which may help integrate object-level structural information.
Conversely, smaller RF size positively correlates with higher texture sensitivity (0.912 in \cref{app_fig:trade-off}c), which aligns with prior studies~\cite{doshi2025visual, brendel2019approximating} showing that models with smaller RF sizes tend to rely on local image features such as textures.

Taken together, these suggest that the RF size of CNNs can explain the observed shape–texture sensitivity trade-off. Although a similar trend appears in the \cue benchmark (\cref{app_fig:trade-off}a), it arises only because the preference metric enforces the shape–texture trade-off by its design. Although these correlations do not strictly imply causality, they highlight the RF as a plausible underlying mechanism that partially explains the observed trade-off.
\newpage

\section{Experimental Details}
\subsection{Models with Different Architectures}
\label{app_sec:model_across_structure}
CNN architectures including BagNet~\cite{brendel2019approximating}, AlexNet~\cite{krizhevsky2012imagenet}, VGG~\cite{simonyan2014very}, ResNet~\cite{he2016deep}, and InceptionV3~\cite{szegedy2016rethinking} are trained using only standard random resized cropping. Vision transformer variants, including ViT~\cite{dosovitskiy2020image}, Swin Transformer~\cite{liu2021Swin}, and CMT~\cite{guo2022cmt}, additionally employ mixed augmentation strategies during training. Due to these differences, comparisons in \cref{fig:figure8} and \cref{fig:figure9} are conducted within each model family for fairness, without comparing across model families.

\begin{table}[H]
    \caption{
    Comparison of models with different architectures in terms of shape/texture sensitivity measured on \ours, preference measured on \cue, and ImageNet-1k top-1 accuracy. SP and TP denote shape and texture preferences, respectively. RF denotes receptive field size. Model colors indicate model families.
    }
    \vspace{-1em}
    \label{app_tbl:model_arch}
    \scriptsize
    \centering
    \setlength{\tabcolsep}{1.3pt}
    \begin{tabular}{clcccccccc}
    \toprule
    & & \multicolumn{1}{l}{} & ImageNet-1k & \multicolumn{2}{c}{REFINED-BIAS} & \multicolumn{4}{c}{Cue-conflict} \\ 
    \cmidrule{4-10} 
    & & \multicolumn{1}{l}{} & & \multicolumn{2}{c}{Full Space} & \multicolumn{2}{c}{Full Space} & \multicolumn{2}{c}{Partial Space} \\ 
    \cmidrule{4-10}
    Arch. & \multicolumn{1}{c}{Models} & RF & Top-1 Acc & Shape-Sens & Texture-Sens & SP & TP & SP & TP \\ 
    \midrule
    \multirow{6}{*}{ViT}
    & \makebox[3mm][l]{\textcolor[rgb]{0.5490, 0.1803, 0.8593}{\scalebox{0.8}{\ding{108}}}} 
    ViT-S [\citenum{dosovitskiy2020image}] 
    & - & 0.7860 & 0.7456 & 0.6671 & 0.2727 & 0.7273 & 0.2890 & 0.7110 \\
    & \makebox[3mm][l]{\textcolor[rgb]{0.5490, 0.1803, 0.8593}{\scalebox{1.0}{\ding{108}}}} 
    ViT-B [\citenum{dosovitskiy2020image}] 
    & - & 0.7898 & 0.7194 & 0.6418 & 0.3385 & 0.6615 & 0.3218 & 0.6782 \\
    & \makebox[3mm][l]{\textcolor[rgb]{0.8593, 0.4666, 0.6}{\scalebox{0.8}{\ding{108}}}} 
    Swin-S [\citenum{liu2021Swin}] 
    & - & 0.8290 & 0.7739 & 0.7505 & 0.2065 & 0.7935 & 0.2397 & 0.7603 \\
    & \makebox[3mm][l]{\textcolor[rgb]{0.8593, 0.4666, 0.6}{\scalebox{1.0}{\ding{108}}}} 
    Swin-B [\citenum{liu2021Swin}] 
    & - & 0.8307 & 0.8047 & 0.7506 & 0.2436 & 0.7564 & 0.2494 & 0.7506 \\
    & \makebox[3mm][l]{\textcolor[rgb]{0.6901, 0.7686, 0.8705}{\scalebox{0.8}{\ding{108}}}} 
    CMT-S [\citenum{guo2022cmt}] 
    & - & 0.8331 & 0.8324 & 0.6085 & 0.3185 & 0.6815 & 0.2859 & 0.7141 \\
    & \makebox[3mm][l]{\textcolor[rgb]{0.6901, 0.7686, 0.8705}{\scalebox{1.0}{\ding{108}}}} 
    CMT-B [\citenum{guo2022cmt}] 
    & - & 0.8452 & 0.8868 & 0.6228 & 0.3312 & 0.6688 & 0.2988 & 0.7012 \\ 
    \midrule
    \multirow{14}{*}{CNN} 
    & \makebox[3mm][l]{\textcolor[rgb]{0.8039, 0.5216, 0.2471}{\scalebox{0.2}{\ding{108}}}} 
    BagNet-9 [\citenum{brendel2019approximating}]
    & 9 & 0.4636 & 0.0517 & 0.7194 & 0.0498 & 0.9502 & 0.0650 & 0.9350 \\
    & \makebox[3mm][l]{\textcolor[rgb]{0.8039, 0.5216, 0.2471}{\scalebox{0.4}{\ding{108}}}} 
    BagNet-17 [\citenum{brendel2019approximating}] 
    & 17 & 0.5865 & 0.0995 & 0.8017 & 0.0076 & 0.9924 & 0.0484 & 0.9516 \\
    & \makebox[3mm][l]{\textcolor[rgb]{0.8039, 0.5216, 0.2471}{\scalebox{0.6}{\ding{108}}}} 
    BagNet-33 [\citenum{brendel2019approximating}] 
    & 33 & 0.6421 & 0.2386 & 0.8008 & 0.0186 & 0.9814 & 0.0622 & 0.9378 \\
    & \makebox[3mm][l]{\textcolor[rgb]{0.6627, 0.6627, 0.6627}{\scalebox{1.0}{\ding{108}}}} 
    AlexNet [\citenum{krizhevsky2012imagenet}] 
    & 167 & 0.5654 & 0.3617 & 0.5648 & 0.2000 & 0.8000 & 0.2636 & 0.7364 \\
    & \makebox[3mm][l]{\textcolor[rgb]{0.9137, 0.5882, 0.4784}{\scalebox{0.2}{\ding{108}}}} 
    VGG-11 [\citenum{simonyan2014very}]  
    & 150 & 0.7037 & 0.4523 & 0.7358 & 0.0753 & 0.9247 & 0.1055 & 0.8945 \\
    & \makebox[3mm][l]{\textcolor[rgb]{0.9137, 0.5882, 0.4784}{\scalebox{0.4}{\ding{108}}}} 
    VGG-13 [\citenum{simonyan2014very}] 
    & 156 & 0.7155 & 0.4337 & 0.7324 & 0.0820 & 0.9180 & 0.1152 & 0.8848 \\
    & \makebox[3mm][l]{\textcolor[rgb]{0.9137, 0.5882, 0.4784}{\scalebox{0.6}{\ding{108}}}} 
    VGG-16 [\citenum{simonyan2014very}] 
    & 212 & 0.7337 & 0.5534 & 0.6672 & 0.0994 & 0.9006 & 0.1129 & 0.8871 \\
    & \makebox[3mm][l]{\textcolor[rgb]{0.9137, 0.5882, 0.4784}{\scalebox{0.8}{\ding{108}}}} 
    VGG-19 [\citenum{simonyan2014very}] 
    & 268 & 0.7422 & 0.6130 & 0.6160 & 0.1373 & 0.8627 & 0.1532 & 0.8468 \\
    & \makebox[3mm][l]{\textcolor[rgb]{0.4157, 0.3529, 0.8039}{\scalebox{0.2}{\ding{108}}}} 
    ResNet-18 [\citenum{he2016deep}] 
    & 435 & 0.6975 & 0.5741 & 0.5674 & 0.1913 & 0.8087 & 0.2367 & 0.7633 \\
    & \makebox[3mm][l]{\textcolor[rgb]{0.4157, 0.3529, 0.8039}{\scalebox{0.4}{\ding{108}}}} 
    ResNet-34 [\citenum{he2016deep}] 
    & 899 & 0.7328 & 0.5680 & 0.5635 & 0.2438 & 0.7562 & 0.2595 & 0.7405 \\
    & \makebox[3mm][l]{\textcolor[rgb]{0.4157, 0.3529, 0.8039}{\scalebox{0.6}{\ding{108}}}} 
    ResNet-50 [\citenum{he2016deep}] 
    & 427 & 0.7613 & 0.5630 & 0.5684 & 0.1944 & 0.8056 & 0.2221 & 0.7779 \\
    & \makebox[3mm][l]{\textcolor[rgb]{0.4157, 0.3529, 0.8039}{\scalebox{0.8}{\ding{108}}}} 
    ResNet-101 [\citenum{he2016deep}] 
    & 971 & 0.7737 & 0.6590 & 0.5303 & 0.2700 & 0.7300 & 0.2866 & 0.7134 \\
    & \makebox[3mm][l]{\textcolor[rgb]{0.4157, 0.3529, 0.8039}{\scalebox{1.0}{\ding{108}}}} 
    ResNet-152 [\citenum{he2016deep}] 
    & 1451 & 0.7831 & 0.6378 & 0.5179 & 0.2612 & 0.7388 & 0.2940 & 0.7060 \\
    & \makebox[3mm][l]{\textcolor[rgb]{0.7412, 0.7176, 0.4196}{\scalebox{1.0}{\ding{108}}}} 
    InceptionV3 [\citenum{szegedy2016rethinking}] 
    & 1311 & 0.7731 & 0.6944 & 0.5233 & 0.5583 & 0.4417 & 0.5038 & 0.4962 \\ 
    \bottomrule
    \end{tabular}
\end{table}
\newpage

\subsection{Models with Different Training Strategies}
\label{app_sec:model_across_train}

\begin{table}[H]
    \caption{
    Effect of training strategies on shape/texture sensitivity measured on \ours, bias measured on \cue, and ImageNet-1k top-1 accuracy. SP and TP denote shape and texture preferences, respectively. RF denotes receptive field size. Model colors indicate model families.
    }
    \vspace{-1em}
    \label{app_tbl4}
    \scriptsize
    \centering
    \setlength{\tabcolsep}{0.6pt}
    \begin{tabular}{lccccccc}
    \cmidrule{1-8}
    & ImageNet-1k & \multicolumn{2}{c}{\ours} & \multicolumn{4}{c}{\Cue}  \\
    \cmidrule{2-8}
    & & \multicolumn{2}{c}{Full Space} & \multicolumn{2}{c}{Full Space} & \multicolumn{2}{c}{Partial Space}\\
    \cmidrule{2-8}
    Models & Top-1 Acc & Shape-Sens & Texture-Sens & SP & TP & SP & TP  \\ 
    \cmidrule{1-8}
    $\blacklozenge$ Vanila ResNet-50 
    & 0.7613 & 0.5630 & 0.5684 & 0.1784 & 0.8216 & 0.2221 & 0.7779 \\ 
    \cmidrule{1-8}
    \makebox[3mm][l]{\textcolor[rgb]{1.0000, 0.8431, 0.0000}{\ding{72}}} A1 [\citenum{wightman2021resnet}] 
    & 0.8009 & 0.8307 & 0.6740 & 0.2193 & 0.7807 & 0.2054 & 0.7946 \\
    \makebox[3mm][l]{\textcolor[rgb]{1.0000, 0.8431, 0.0000}{\scalebox{0.3}{\BlackKnightOnWhite}}} A2 [\citenum{wightman2021resnet}] 
    & 0.7978 & 0.7515 & 0.6880 & 0.1617 & 0.8383 & 0.1908 & 0.8092 \\
    \makebox[3mm][l]{\textcolor[rgb]{1.0000, 0.8431, 0.0000}{\scalebox{0.3}{\BlackRookOnWhite}}} A3 [\citenum{wightman2021resnet}] 
    & 0.7754 & 0.7737 & 0.6927 & 0.1420 & 0.8580 & 0.1864 & 0.8136 \\
    \makebox[3mm][l]{\textcolor[rgb]{1.0000, 0.8431, 0.0000}{\ding{67}}} B [\citenum{wightman2021resnet}]  
    & 0.7925 & 0.8367 & 0.5960 & 0.2014 & 0.7986 & 0.2011 & 0.7989 \\
    \makebox[3mm][l]{\textcolor[rgb]{1.0000, 0.8431, 0.0000}{\scalebox{1.3}{\ding{171}}}} C1 [\citenum{wightman2021resnet}] 
    & 0.7974 & 0.8532 & 0.6782 & 0.2230 & 0.7770 & 0.1979 & 0.8021 \\
    \makebox[3mm][l]{\textcolor[rgb]{1.0000, 0.8431, 0.0000}{\ding{115}}} C2 [\citenum{wightman2021resnet}] 
    & 0.7990 & 0.8382 & 0.6607 & 0.2143 & 0.7857 & 0.2171 & 0.7829 \\
    \makebox[3mm][l]{\textcolor[rgb]{1.0000, 0.8431, 0.0000}{\ding{168}}} D [\citenum{wightman2021resnet}]  
    & 0.7988 & 0.8457 & 0.6363 & 0.1958 & 0.8042 & 0.1957 & 0.8043 \\
    \makebox[3mm][l]{\textcolor[rgb]{1.0000, 0.8431, 0.0000}{\scalebox{0.3}{\BlackKingOnWhite}}} V2 [\citenum{wightman2021resnet}] 
    & 0.8033 & 0.7467 & 0.7386 & 0.2116 & 0.7884 & 0.1874 & 0.8126 \\ 
    \cmidrule{1-8}
    \makebox[3mm][l]{\textcolor[rgb]{0.757, 0.490, 0.922}{\scalebox{0.3}{\BlackKnightOnWhite}}} AugMix [\citenum{hendrycks2019augmix}]
    & 0.7752 & 0.7285 & 0.5278 & 0.2776 & 0.7224 & 0.3177 & 0.6823 \\
    \makebox[3mm][l]{\textcolor[rgb]{0.757, 0.490, 0.922}{\scalebox{0.3}{\BlackBishopOnWhite}}} PixMix [\citenum{hendrycks2022pixmix}]    
    & 0.7807 & 0.6648 & 0.5956 & 0.2796 & 0.7204 & 0.3196 & 0.6804 \\
    \makebox[3mm][l]{\textcolor[rgb]{0.757, 0.490, 0.922}{$\blacklozenge$}} OpticsAugment [\citenum{muller2023classification}]
    & 0.7421 & 0.6246 & 0.5493 & 0.1953 & 0.8047 & 0.2652 & 0.7348 \\
    \makebox[3mm][l]{\textcolor[rgb]{0.757, 0.490, 0.922}{\scalebox{1.3}{\ding{171}}}} DeepAugment [\citenum{hendrycks2021many}]         
    & 0.7664 & 0.7067 & 0.5225 & 0.3372 & 0.6628 & 0.3557 & 0.6443 \\
    \makebox[3mm][l]{\textcolor[rgb]{0.757, 0.490, 0.922}{\ding{168}}} PRIME [\citenum{modas2022prime}]                  
    & 0.7689 & 0.7995 & 0.5293 & 0.3222 & 0.6778 & 0.3378 & 0.6622 \\ 
    \cmidrule{1-8}
    \makebox[3mm][l]{\textcolor[rgb]{0.329, 0.518, 0.922}{\scalebox{0.3}{\BlackRookOnWhite}} } ShapeNet: SIN+IN [\citenum{geirhos2018imagenet}]    
    & 0.7458 & 0.7251 & 0.5137 & 0.3717 & 0.6283 & 0.3657 & 0.6343 \\
    \makebox[3mm][l]{\textcolor[rgb]{0.329, 0.518, 0.922}{\ding{67}}} ShapeNet: SIN+IN+FT [\citenum{geirhos2018imagenet}] 
    & 0.7671 & 0.6798 & 0.5365 & 0.1993 & 0.8007 & 0.2290 & 0.7710 \\
    \makebox[3mm][l]{\textcolor[rgb]{0.329, 0.518, 0.922}{\scalebox{1.3}{\ding{171}}}} Shape Bias Augmentation [\citenum{li2020shape}]     
    & 0.7619 & 0.6843 & 0.5463 & 0.3000 & 0.7000 & 0.3128 & 0.6872 \\
    \cmidrule{1-8}
    \makebox[3mm][l]{\textcolor[rgb]{0.133, 0.545, 0.133}{\scalebox{1.3}{\ding{171}}}} MoCo v3 [\citenum{chen2021empirical}]    
    & 0.7458 & 0.6111 & 0.4409 & 0.2704 & 0.7296 & 0.3093 & 0.6907 \\
    \makebox[3mm][l]{\textcolor[rgb]{0.133, 0.545, 0.133}{\scalebox{0.3}{\BlackRookOnWhite}}} SimCLR v2 [\citenum{chen2020big}]                 
    & 0.7489 & 0.5869 & 0.4303 & 0.2043 & 0.7957 & 0.2528 & 0.7472  \\
    \makebox[3mm][l]{\textcolor[rgb]{0.133, 0.545, 0.133}{\ding{67}}} DINO v1 [\citenum{caron2021emerging}]             
    & 0.7527 & 0.6705 & 0.5314 & 0.1136 & 0.8864 & 0.1639 & 0.8361  \\
    \cmidrule{1-8}
    \makebox[3mm][l]{\textcolor[rgb]{0.922, 0.537, 0.173}{\scalebox{0.4}{\ding{108}}}} PGD-AT ($l_2$, $\epsilon=0.05$) [\citenum{madry2017towards, salman2020adversarially}] 
    & 0.7557 & 0.6645 & 0.5468 & 0.1878 & 0.8122 & 0.2678 & 0.7322  \\
    \makebox[3mm][l]{\textcolor[rgb]{0.922, 0.537, 0.173}{\scalebox{0.5}{\ding{108}}}} PGD-AT ($l_2$, $\epsilon=0.1$) [\citenum{madry2017towards, salman2020adversarially}] 
    & 0.7478 & 0.5070 & 0.5484 & 0.2436 & 0.7564 & 0.3198 & 0.6802 \\
    \makebox[3mm][l]{\textcolor[rgb]{0.922, 0.537, 0.173}{\scalebox{0.6}{\ding{108}}}} PGD-AT ($l_2$, $\epsilon=0.25$) [\citenum{madry2017towards, salman2020adversarially}] 
    & 0.7412 & 0.5703 & 0.5311 & 0.2941 & 0.7059 & 0.3786 & 0.6214 \\
    \makebox[3mm][l]{\textcolor[rgb]{0.922, 0.537, 0.173}{\scalebox{0.7}{\ding{108}}}} PGD-AT ($l_2$, $\epsilon=0.5$) [\citenum{madry2017towards, salman2020adversarially}] 
    & 0.7317 & 0.5070 & 0.5481 & 0.4680 & 0.5320 & 0.4329 & 0.5671 \\
    \makebox[3mm][l]{\textcolor[rgb]{0.922, 0.537, 0.173}{\scalebox{0.8}{\ding{108}}}} PGD-AT ($l_2$, $\epsilon=1$) [\citenum{madry2017towards, salman2020adversarially}] 
    & 0.7042 & 0.5363 & 0.5406 & 0.5771 & 0.4229 & 0.5306 & 0.4694 \\
    \makebox[3mm][l]{\textcolor[rgb]{0.922, 0.537, 0.173}{\scalebox{0.9}{\ding{108}}}} PGD-AT ($l_2$, $\epsilon=3$) [\citenum{madry2017towards, salman2020adversarially}] 
    & 0.6283 & 0.4918 & 0.4978 & 0.8615 & 0.1385 & 0.6883 & 0.3117  \\
    \makebox[3mm][l]{\textcolor[rgb]{0.922, 0.537, 0.173}{\scalebox{1.0}{\ding{108}}}} PGD-AT ($l_2$, $\epsilon=5$) [\citenum{madry2017towards, salman2020adversarially}] 
    & 0.5613 & 0.4292 & 0.5027 & 0.9029 & 0.0971 & 0.7379 & 0.2621 \\
    \makebox[3mm][l]{\textcolor[rgb]{0.922, 0.537, 0.173}{\scalebox{0.2}{\ding{110}}}} PGD-AT ($l_\infty$, $\epsilon=0.5$) [\citenum{madry2017towards, salman2020adversarially}] 
    & 0.7317 & 0.6021 & 0.5383 & 0.3864 & 0.6136 & 0.4124 & 0.5876 \\
    \makebox[3mm][l]{\textcolor[rgb]{0.922, 0.537, 0.173}{\scalebox{0.4}{\ding{110}}}} PGD-AT ($l_\infty$, $\epsilon=1$) [\citenum{madry2017towards, salman2020adversarially}] 
    & 0.7042 & 0.5583 & 0.5384 & 0.5571 & 0.0229 & 0.4967 & 0.5033 \\
    \makebox[3mm][l]{\textcolor[rgb]{0.922, 0.537, 0.173}{\scalebox{0.6}{\ding{110}}}} PGD-AT ($l_\infty$, $\epsilon=2$) [\citenum{madry2017towards, salman2020adversarially}] 
    & 0.6908 & 0.5237 & 0.5086 & 0.6946 & 0.3054 & 0.5756 & 0.4244 \\
    \makebox[3mm][l]{\textcolor[rgb]{0.922, 0.537, 0.173}{\scalebox{0.8}{\ding{110}}}} PGD-AT ($l_\infty$, $\epsilon=4$) [\citenum{madry2017towards, salman2020adversarially}] 
    & 0.6386 & 0.4963 & 0.5144 & 0.8302 & 0.1698 & 0.6429 & 0.3571 \\
    \makebox[3mm][l]{\textcolor[rgb]{0.922, 0.537, 0.173}{\scalebox{1.0}{\ding{110}}}} PGD-AT ($l_\infty$, $\epsilon=8$) [\citenum{madry2017towards, salman2020adversarially}] 
    & 0.5452 & 0.5268 & 0.5039 & 0.9087 & 0.0913 & 0.7361 & 0.2639 \\
    \cmidrule{1-8}
    \end{tabular}
\end{table}
\clearpage

\subsection{Confidence Intervals for Models with Different Training Strategies}
\begin{table}[H]
    \caption{
    Confidence intervals ($95\%$ CI) of shape and texture sensitivities measured on \ours dataset for models trained with different training strategies. The colors indicate model families. Overlapping CIs suggest similar shape or texture sensitivities, whereas non-overlapping CIs suggest statistically significant differences.
    }
    \vspace{-1em}
    \scriptsize
    \centering
    \setlength{\tabcolsep}{6.9pt}
    \begin{tabular}{lcccccc}
    \cmidrule{1-7}
     & \multicolumn{3}{c}{Shape-Sens} & \multicolumn{3}{c}{Texture-Sens} \\
    \cmidrule{1-7}
     Models 
       & \shortstack{Lower} 
       & Mean 
       & \shortstack{Upper} 
       & \shortstack{Lower} 
       & Mean 
       & \shortstack{Upper} \\
    \cmidrule{1-7}
    $\blacklozenge$ Vanila ResNet-50 
    & 0.5548 &  0.5637 & 0.5724 
    & 0.5538 & 0.5690 & 0.5823 
    \\
    \cmidrule{1-7}
    \makebox[3mm][l]{\textcolor[rgb]{1.0000, 0.8431, 0.0000}{\ding{72}}} A1 [\citenum{wightman2021resnet}] 
    & 0.8215 & 0.8301 & 0.8405 
    & 0.6532 & 0.6749 & 0.6839 
    \\
    \makebox[3mm][l]{\textcolor[rgb]{1.0000, 0.8431, 0.0000}{\scalebox{0.3}{\BlackKnightOnWhite}}} A2 [\citenum{wightman2021resnet}] 
    &  0.7402 & 0.7506 & 0.7593 
    & 0.6721 & 0.6883 & 0.7013 
    \\
    \makebox[3mm][l]{\textcolor[rgb]{1.0000, 0.8431, 0.0000}{\scalebox{0.3}{\BlackRookOnWhite}}} A3 [\citenum{wightman2021resnet}] 
    & 0.7270 & 0.7364 & 0.7462
    & 0.6715 & 0.6931 & 0.7085 
    \\
    \makebox[3mm][l]{\textcolor[rgb]{1.0000, 0.8431, 0.0000}{\ding{67}}} B [\citenum{wightman2021resnet}]  
    & 0.8283 & 0.8357 & 0.8433  & 0.5764 & 0.5961 & 0.6149 
    \\
    \makebox[3mm][l]{\textcolor[rgb]{1.0000, 0.8431, 0.0000}{\scalebox{1.3}{\ding{171}}}} C1 [\citenum{wightman2021resnet}] 
    & 0.8440 & 0.8539 & 0.8658  & 0.6627 & 0.6793 & 0.6948 
    \\
    \makebox[3mm][l]{\textcolor[rgb]{1.0000, 0.8431, 0.0000}{\ding{115}}} C2 [\citenum{wightman2021resnet}] 
    & 0.8300 & 0.8353 & 0.8445 & 0.6463 & 0.6624 & 0.6745 
    \\
    \makebox[3mm][l]{\textcolor[rgb]{1.0000, 0.8431, 0.0000}{\ding{168}}} D [\citenum{wightman2021resnet}] 
    &  0.8380 & 0.8459  & 0.8542 & 0.6202 & 0.6376 & 0.6510 
    \\
    \makebox[3mm][l]{\textcolor[rgb]{1.0000, 0.8431, 0.0000}{\scalebox{0.3}{\BlackKingOnWhite}}} V2 [\citenum{wightman2021resnet}] 
    & 0.7352  & 0.7440  & 0.7538 & 0.7249 & 0.7402 & 0.7507 
    \\ 
    \cmidrule{1-7}
    \makebox[3mm][l]{\textcolor[rgb]{0.757, 0.490, 0.922}{\ding{118}}} AugMix [\citenum{hendrycks2019augmix}] 
    & 0.7209 & 0.7299 & 0.7391
    & 0.5070 & 0.5271 & 0.5417 
    \\ 
    \makebox[3mm][l]{\textcolor[rgb]{0.757, 0.490, 0.922}{\scalebox{0.3}{\BlackBishopOnWhite}}} PixMix [\citenum{hendrycks2022pixmix}] 
    & 0.6600 & 0.6654 & 0.6740
    & 0.5746 & 0.5956 & 0.6104
    \\ 
    \makebox[3mm][l]{\textcolor[rgb]{0.757, 0.490, 0.922}{$\blacklozenge$}} OpticsAugment [\citenum{muller2023classification}] 
    & 0.6209 & 0.6276 & 0.6376 
    & 0.5263 & 0.5475 & 0.5633 
    \\ 
    \makebox[3mm][l]{\textcolor[rgb]{0.757, 0.490, 0.922}{\scalebox{1.3}{\ding{171}}}} DeepAugment [\citenum{hendrycks2021many}] 
    & 0.6980 & 0.7063 & 0.7174 
    & 0.5139 & 0.5242 & 0.5339 
    \\ 
    \makebox[3mm][l]{\textcolor[rgb]{0.757, 0.490, 0.922}{\ding{168}}} PRIME [\citenum{modas2022prime}] 
    & 0.7929 & 0.8019 & 0.8130 
    & 0.5124 & 0.5320 & 0.5466 
    \\ 
    \cmidrule{1-7}
    \makebox[3mm][l]{\textcolor[rgb]{0.329, 0.518, 0.922}{\scalebox{0.3}{\BlackRookOnWhite}} } ShapeNet: SIN+IN [\citenum{geirhos2018imagenet}] 
    & 0.7165 & 0.7264 & 0.7350 
    & 0.4968 & 0.5134 & 0.5263 
    \\
    \makebox[3mm][l]{\textcolor[rgb]{0.329, 0.518, 0.922}{\ding{67}}} ShapeNet: SIN+IN+FT [\citenum{geirhos2018imagenet}] 
    & 0.6744 & 0.6819 & 0.6931 
    & 0.5167 & 0.5360 & 0.5507 
    \\
    \makebox[3mm][l]{\textcolor[rgb]{0.329, 0.518, 0.922}{\scalebox{1.3}{\ding{171}}}} Shape Bias Augmentation [\citenum{li2020shape}] 
    & 0.6743 & 0.6850 & 0.6971 
    & 0.5283 & 0.5453 & 0.5599 
    \\
    \cmidrule{1-7}
    \makebox[3mm][l]{\textcolor[rgb]{0.133, 0.545, 0.133}{\scalebox{1.3}{\ding{171}}}} MoCo v3 [\citenum{chen2021empirical}] 
    & 0.6042 & 0.6134 & 0.6231 
    & 0.4266 & 0.4429 & 0.4568 
    \\
    \makebox[3mm][l]{\textcolor[rgb]{0.133, 0.545, 0.133}{\scalebox{0.3}{\BlackRookOnWhite}}} SimCLR v2 [\citenum{chen2020big}] 
    &  0.5814 & 0.5878 & 0.5993 
    & 0.4152 & 0.4338 & 0.4477 
    \\
    \makebox[3mm][l]{\textcolor[rgb]{0.133, 0.545, 0.133}{\ding{67}}} DINO v1 [\citenum{caron2021emerging}] 
    & 0.6619 & 0.6684 & 0.6781 
    & 0.5205 & 0.5333 & 0.5474 
    \\
    \cmidrule{1-7}
    \makebox[3mm][l]{\textcolor[rgb]{0.922, 0.537, 0.173}{\scalebox{0.4}{\ding{108}}}} PGD-AT ($l_2$, $\epsilon=0.05$) [\citenum{madry2017towards, salman2020adversarially}] 
    & 0.6564 & 0.6681 & 0.6780
    & 0.5321 & 0.5495 & 0.5653 
    \\
    \makebox[3mm][l]{\textcolor[rgb]{0.922, 0.537, 0.173}{\scalebox{0.5}{\ding{108}}}} PGD-AT ($l_2$, $\epsilon=0.1$) [\citenum{madry2017towards, salman2020adversarially}]
    & 0.5022 & 0.5088 & 0.5161
    & 0.5312 & 0.5496 & 0.5625 
    \\
    \makebox[3mm][l]{\textcolor[rgb]{0.922, 0.537, 0.173}{\scalebox{0.6}{\ding{108}}}} PGD-AT ($l_2$, $\epsilon=0.25$) [\citenum{madry2017towards, salman2020adversarially}] 
    & 0.5664 & 0.5743 & 0.5816
    & 0.5210 & 0.5329 & 0.5428 
    \\
    \makebox[3mm][l]{\textcolor[rgb]{0.922, 0.537, 0.173}{\scalebox{0.7}{\ding{108}}}} PGD-AT ($l_2$, $\epsilon=0.5$) [\citenum{madry2017towards, salman2020adversarially}] 
    & 0.5096 & 0.5149 & 0.5213
    & 0.5353 & 0.5477 & 0.5604 
    \\
    \makebox[3mm][l]{\textcolor[rgb]{0.922, 0.537, 0.173}{\scalebox{0.8}{\ding{108}}}} PGD-AT ($l_2$, $\epsilon=1$) [\citenum{madry2017towards, salman2020adversarially}] 
    & 0.5281 & 0.5383  & 0.5493
    & 0.5321 & 0.5423 & 0.5556 
    \\
    \makebox[3mm][l]{\textcolor[rgb]{0.922, 0.537, 0.173}{\scalebox{0.9}{\ding{108}}}} PGD-AT ($l_2$, $\epsilon=3$) [\citenum{madry2017towards, salman2020adversarially}] 
    & 0.4881 & 0.4955 & 0.5021
    & 0.4899 & 0.5010 & 0.5143 
    \\
    \makebox[3mm][l]{\textcolor[rgb]{0.922, 0.537, 0.173}{\scalebox{1.0}{\ding{108}}}} PGD-AT ($l_2$, $\epsilon=5$) [\citenum{madry2017towards, salman2020adversarially}] 
    & 0.4249 & 0.4315 & 0.4402
    & 0.4947 & 0.5056 & 0.5193 
    \\
    \makebox[3mm][l]{\textcolor[rgb]{0.922, 0.537, 0.173}{\scalebox{0.2}{\ding{110}}}} PGD-AT ($l_\infty$, $\epsilon=0.5$) [\citenum{madry2017towards, salman2020adversarially}] 
    & 0.5992 & 0.6065 & 0.6125
    & 0.5263 & 0.5394 & 0.5526 
    \\
    \makebox[3mm][l]{\textcolor[rgb]{0.922, 0.537, 0.173}{\scalebox{0.4}{\ding{110}}}} PGD-AT ($l_\infty$, $\epsilon=1$) [\citenum{madry2017towards, salman2020adversarially}] 
    & 0.5542 & 0.5614  & 0.5707
    & 0.5249 & 0.5387 & 0.5561 
    \\
    \makebox[3mm][l]{\textcolor[rgb]{0.922, 0.537, 0.173}{\scalebox{0.6}{\ding{110}}}} PGD-AT ($l_\infty$, $\epsilon=2$) [\citenum{madry2017towards, salman2020adversarially}] 
    & 0.5231 & 0.5279 & 0.5370
    & 0.4986 & 0.5108 & 0.5229 
    \\
    \makebox[3mm][l]{\textcolor[rgb]{0.922, 0.537, 0.173}{\scalebox{0.8}{\ding{110}}}} PGD-AT ($l_\infty$, $\epsilon=4$) [\citenum{madry2017towards, salman2020adversarially}] 
    & 0.4910 & 0.5013 & 0.5117
    & 0.5042 & 0.5142 & 0.5277 
    \\
    \makebox[3mm][l]{\textcolor[rgb]{0.922, 0.537, 0.173}{\scalebox{1.0}{\ding{110}}}} PGD-AT ($l_\infty$, $\epsilon=8$) [\citenum{madry2017towards, salman2020adversarially}] 
    & 0.5216 & 0.5312 & 0.5381
    & 0.4924 & 0.5038 & 0.5160 
    \\
    \cmidrule{1-7}
    \end{tabular}
\end{table}
\clearpage

\subsection{Configurations of Models Trained with Mixed Augmentations}
\label{app_sec:timm_model_details}
\begin{table}[h]
    \label{App:timm_mixture_aug}
    \caption{
    Training settings and augmentation combinations of models from timm~\cite{wightman2021resnet}. AGC denotes adaptive gradient clipping, while CE and BCE refer to cross-entropy and binary cross-entropy losses, respectively.
    }
    \vspace{-1em}
    \label{App:timm_train_setting}
    \scriptsize
    \centering
    \setlength{\tabcolsep}{6.1pt}
    \begin{tabular}{lcccccccc}
    \toprule
    Augmentations         & A1     & A2     & A3     & B       & C1     & C2     & D      & V2     \\
    \midrule
    RandAug               & \cellcolor[HTML]{ffb4ae}O      & \cellcolor[HTML]{efcee2}O      & \cellcolor[HTML]{ffb4ae}O      & \cellcolor[HTML]{efcee2}O       & \cellcolor[HTML]{ffb4ae}O      & \cellcolor[HTML]{efcee2}O      & \cellcolor[HTML]{ffb4ae}O      & X      \\
    MixUp                 & \cellcolor[HTML]{ffb4ae}O      & \cellcolor[HTML]{efcee2}O      & \cellcolor[HTML]{ffb4ae}O      & \cellcolor[HTML]{efcee2}O       & \cellcolor[HTML]{ffb4ae}O      & \cellcolor[HTML]{efcee2}O      & \cellcolor[HTML]{ffb4ae}O      & \cellcolor[HTML]{efcee2}O      \\
    CutMix                & \cellcolor[HTML]{ffb4ae}O      & \cellcolor[HTML]{efcee2}O      & \cellcolor[HTML]{ffb4ae}O      & X       & \cellcolor[HTML]{ffb4ae}O      & \cellcolor[HTML]{efcee2}O      & \cellcolor[HTML]{ffb4ae}O      & \cellcolor[HTML]{efcee2}O      \\
    Random Erasing        & X      & X      & X      & \cellcolor[HTML]{efcee2}O       & \cellcolor[HTML]{ffb4ae}O      & \cellcolor[HTML]{efcee2}O      & \cellcolor[HTML]{ffb4ae}O      & \cellcolor[HTML]{efcee2}O      \\
    Repeated Aug. & \cellcolor[HTML]{ffb4ae}O      & \cellcolor[HTML]{efcee2}O      & X      & X       & X      & \cellcolor[HTML]{efcee2}O      & X      & \cellcolor[HTML]{efcee2}O      \\
    Auto Aug.     & X      & X      & X      & X       & X      & X      & X      & \cellcolor[HTML]{efcee2}O      \\
    Label Smoothing       & \cellcolor[HTML]{ffb4ae}O      & X      & X      & \cellcolor[HTML]{efcee2}O       & \cellcolor[HTML]{ffb4ae}O      & \cellcolor[HTML]{efcee2}O      & \cellcolor[HTML]{ffb4ae}O      & \cellcolor[HTML]{efcee2}O      \\
    \midrule
    EMA                   & X      & X      & X      & O       & X      & X      & X      & O      \\
    Grad Clipping         & X      & X      & X      & X       & AGC    & AGC    & X      & X      \\
    Optimizer             & lamb   & lamb   & lamb   & rmsprop & sgd    & sgd    & adamp  & sgd    \\
    LR decay              & cosine & cosine & cosine & step    & cosine & cosine & cosine & cosine \\
    Loss                  & BCE    & BCE    & BCE    & CE      & CE     & CE     & BCE    & CE     \\
    Epoch                 & 600    & 300    & 100    & 600     & 800    & 800    & 600    & 600    \\
    Weight Decay          & 0.01   & 0.02   & 0.02   & 7e-6    & 1e-5   & 1e-5   & 0.01   & 2e-5   \\
    Stochastic Depth      & 0.05   & 0.05   & X      & 0.1     & 0.1    & 0.1    & 0.05   & X      \\
    \bottomrule
    \end{tabular}
\end{table}

\end{document}